\documentclass{article}

\usepackage{arxiv}

\usepackage[utf8]{inputenc} 
\usepackage[T1]{fontenc}    
\usepackage{hyperref}       
\usepackage{url}            
\usepackage{booktabs}       
\usepackage{amsfonts}       
\usepackage{nicefrac}       
\usepackage{microtype}      
\usepackage{lipsum}		
\usepackage{graphicx}  
\usepackage{natbib}
\usepackage{doi}
\usepackage{appendix} 

\usepackage{cognac}
\usepackage{graphicx}
\usepackage{multirow} 
\usepackage{nameref} 

\newlength{\sssep}
\newcommand{\sss}[3]{#1\hspace{\sssep}/\hspace{\sssep}#2\hspace{\sssep}/\hspace{\sssep}#3}
\newcommand{\rpf}[3]{0.#3 (0.#2 / 0.#1 )}
\newcommand{\rpff}[3]{0.#3 (#2.0 / 0.#1 )}
\newcommand{\rpfb}[3]{\textbf{0.#3} (0.#2 / 0.#1)}
\newcommand{\motif}[2]{#1, \textit{#2}}

\title{Automated Motif Indexing on the Arabian Nights}

\author{ \href{https://orcid.org/0000-0001-8491-8780}{\includegraphics[scale=0.06]{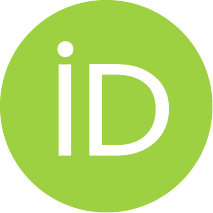}\hspace{1mm}Ibrahim H. Alyami}$^{1,2}$\\
	$^{1}$College of Computer Science and Information Systems\\
	Najran University\\
	King Abdul Aziz Rd, 66462, Najran, KSA \\
	\texttt{ihalmerdef@nu.edu.sa} \\
	\And
	\href{https://orcid.org/0000-0001-5716-9526}{\includegraphics[scale=0.06]{orcid.pdf}\hspace{1mm}Mark A. Finlayson}$^{2}$ \\
	$^{2}$Knight Foundation School of Computing and Information Sciences\\
	Florida International University\\
	11200 SW 8th Street, 33199, FL, USA \\
	\texttt{markaf@fiu.edu} \\
}

\renewcommand{\shorttitle}{\textit{arXiv} Template}

\begin{document}
\maketitle

\begin{abstract}
Motifs are non-commonplace, recurring narrative elements, often found originally in folk stories. In addition to being of interest to folklorists, motifs appear as metaphoric devices in modern news, literature, propaganda, and other cultural texts. Finding expressions of motifs in the original folkloristic text is useful for both folkloristic analysis (motif indexing) as well as for understanding the modern usage of motifs (motif detection and interpretation). Prior work has primarily shown how difficult these problems are to tackle using automated techniques. We present the first computational approach to motif indexing. Our choice of data is a key enabler: we use a large, widely available text (the \textit{Arabian Nights}) paired with a detailed motif index (by El-Shamy in 2006), which overcomes the common problem of inaccessibility of texts referred to by the index. We created a manually annotated corpus that identified 2,670 motif expressions of 200 different motifs across 58,450 sentences for training and testing. We tested five types of approaches for detecting motif expressions given a motif index entry: (1) classic retrieve and re-rank using keywords and a fine-tuned cross-encoder; (2) off-the-shelf embedding models; (3) fine-tuned embedding models; (4) generative prompting of off-the-shelf LLMs in $N$-shot setups; and (5) the same generative approaches on LLMs fine-tuned with LoRA. Our best performing system is a fine-tuned Llama3 model which achieves an overall performance of 0.85 $F_1$. 
\end{abstract}


\section{Introduction}
Motifs are non-commonplace, specific, recurring narrative elements that are often found originally in folk stories, and are more generally deployed in culturally inflected materials. Motifs are interesting because they are a compact source of cultural knowledge, in that many motifs concisely communicate a constellation of related ideas, associations, and assumptions. For example, ``troll under a bridge'' is a motif common in Western cultures with roots in Scandinavia. To those familiar with the motif, it entails a number of related ideas that are not directly communicated by the surface meaning of the words: the bridge is along the critical path of the hero, and he must cross it to achieve his goal; the troll often lives under the bridge, crawling out to waylay innocent passers-by; the troll charges a toll or demands something for crossing the bridge; the troll is a squatter, not the officially sanctioned master of the bridge; the troll enforces his illegitimate claim through threat of physical violence; and the hero often ends up battling (and defeating) the troll instead of paying the toll.

While motifs usually originate in folkloristic material, they are frequently used in modern discourse, and can be easily found in speeches, news reports, press releases, propaganda, books, movies, and any type of language where cultural knowledge is deployed. An excellent example of such modern usage is the Islamist motif \textit{Pharaoh}. The Pharaoh appears in stories found in the Hebrew and Christian Bibles and the Qur’an; in those stories, the Pharaoh comes into conflict with Moses and his attempts to free the Hebrews from Egyptian slavery. The Pharaoh is an arrogant and obstinate tyrant who defies the will of God and is punished for it. In modern Islamist extremist narratives, the Pharaoh is a symbol of struggles against anti-Islamic regimes and has been invoked against leaders such as Anwar Sadat of Egypt, Ariel Sharon of Israel, and George W. Bush of the United States, the last of whom Osama bin Laden referred to as the ``pharaoh of the century'' \citep{halverson2011master}. Further, one must be familiar with Islamic folkloristic tradition to understand the use of this motif in modern language; its meaning is metaphoric and obscure to those not versed in the tradition of the group.

Automatically detecting and interpreting motif expressions in modern language is a challenging problem, as has been demonstrated in other work \citep{yarlott2016learning,yarlott2022finding,yarlottto2022communicating,yarlott2024golem, acharya2022integrating,acharya2024discovering}. Prior work defined the task of \textit{motif detection}, which is finding a motif expression in non-folkloristic materials. In that task, even a word-by-word expression of a motif must be differentiated into \textsc{motific}, \textsc{eponymic}, \textsc{referential}, or \textsc{unrelated} types~\citep{yarlott2024golem}, where only the \textsc{motific} usage is intended to call to mind the implicit associations of the motif. Here we tackle a different task, which is detecting appearances of a motif in the original folkloristic materials, which we will call \textit{motif indexing}. In this task, the use of a motif in the original folkloristic materials is by definition an expression of the motif, so motif expressions do not need to be divided into subtypes.  The motif indexing task requires access to a motif index that lists the motifs used in the text.  Methods that address this task would allow the automatic mining of positive and negative examples to train and test other stages of motif understanding, such as discovery of novel motifs \cite[i.e., \textit{motif discovery}:][]{yarlott2016learning}, identification of motif usage in non-folkloristic materials~\citep[i.e., \textit{motif detection}, as above:][]{yarlott2022finding, yarlott2024golem}, and interpretation of the meaning of motific language~\citep[i.e., \textit{motif interpretation}:][]{acharya2022integrating,acharya2024discovering}.

One barrier to developing systems for automated motif indexing has been data availability. In particular, it is hard to assemble a large set of actual folkloristic texts indexed by folklorists. While folklorists have assembled many motif indices for particular groups or countries, often reading and analyzing hundreds of stories to identify their motific patterns. They have usually worked from large collections of folkloristic material that have not been digitized, or only digitized in part. Moreover many indices were assembled quite some time ago (the height of motif index construction was the first half of the 20th century) and the original texts are no longer easily accessible. Therefore, while there are many motif indices, the texts to accompany them are usually not readily available. We solve this problem by leveraging a specific motif index---that written by~\cite{ElShamy:06}---which catalogs the motifs found in the well-known text \textit{The Arabian Nights}. This index is unique in that it is both large ($\sim$5,000 motifs) and indexes a single text (1,398,863 tokens, 45,769 sentences) that is readily available in modern translation~\citep{irwin2010arabian}. These data enable systematic development of automatic motif indexing techniques, which we describe in this article. 

This paper is structured as follows. We first provide background on motifs and review related computational work (\nameref{sec:relatedwork}). We then describe \textit{The Arabian Nights} index and text, and describe pre-processing (\nameref{sec:data}). Next we analyze the complexity of motif structure and usage, which allows us a more nuanced view during evaluation (\nameref{sec:analysis}). We then describe the data annotation in detail, covering the annotator training, annotation procedure, and the annotations collected (\nameref{sec:annotation}). We describe the methods we explored, including classic retrieve and re-rank, off-the-shelf and fine-tuned embedding models, and generative prompting of off-the-shelf and fine-tuned LLMs in $N$-shot setups. (\nameref{sec:methods}). Finally, we discuss the results and possible next steps (\nameref{sec:discussion}) and we conclude with a list of our contributions (\nameref{sec:contributions}).

\section{Related Work and Background}
\label{sec:relatedwork}
\subsection{Motifs in Folkloristics}

Folklorists have proposed several definitions \textit{motif}. Stith Thompson, an American folklorist, provided a now-classic definition in his book \textit{The Folktale}, where he defined a motif as:

\begin{quote}\ldots the smallest element in a tale having a power to persist in tradition. In order to have this power it must have something unusual and striking about it. Most motifs fall into three classes. First are the actors in a tale – gods, or unusual animals, or marvelous creatures like witches, ogres, or fairies, or even conventionalized human characters like the favorite youngest child or the cruel step-mother. Second come certain items in the background of the action – magic objects, unusual customs, strange beliefs, and the like. In the third place there are single incidents – and these comprise the great majority of motifs~\cite[p. 415-416]{thompson1977folktale}.
\end{quote}

Another definition is found in a dictionary of literary terms: 

\begin{quote}
Motif: A theme, character, or verbal pattern which recurs in literature or folklore. \textit{The reveler who blasphemes upon a grave and is later dragged to damnation by the ghost of the man who was buried there} is a widespread folklore motif which later becomes part of the \textit{Don Juan} legend. A motif may be a theme which runs through a number of different works. The motif of the imperishability of art, for example, appears in Shakespeare, Keats, Yeats, and many other writers. A recurring element within a single work is also called a motif. Among the many motifs that appear and reappear in Joyce's Ulysses, for example, are \textit{Plumtree's Potted Meat}, \textit{the man in the brown mackintosh}, and \textit{the one-legged sailor}.~\cite[p. 129]{BecksonGanz:60}.
\end{quote}

\citet{freedman1971literary} claims that five factors establish a motif:
\begin{itemize}
\item \textbf{Frequency:} Motifs should occur fairly often in a corpus of tales. For example, the \textit{magic carpet} motif represents an object that is used by the hero to fly from one place to another place in a magically faster way. This is a motif that is repeated frequently in key events in certain parts of the Arabian Nights stories.

\item \textbf{Unavoidable and Unlikely:} The \textit{magic carpet} represents freedom and adventure themes
and is important to represent these themes. Therefore, the motif cannot be avoided. It is ``unlikely'' because it carries unexpected meanings beyond its physical presence. A motif may stand out more to the audience and have more effect if it is unexpected or rare.

\item \textbf{Contextually Significant:} Motifs that appear in main events are more important than motifs that appear in non-main events.

\item \textbf{Relevant and Coherent:} All expressions of the motif should work to serve the primary
meaning behind that motif. The more united and related to the motif’s components are, the more effective they are, and vice versa.

\item \textbf{Symbolic Appropriateness:} Motifs should represent the theme correctly. Next, we highlight the differences between motifs and themes.
\end{itemize}

\citet{abbott2021cambridge} contrasted motifs and the related idea of a \textit{theme}.

\begin{quote}
Motif: A discrete thing, image, or phrase that is repeated in a narrative. Theme, by contrast, is a more generalized or abstract concept that is suggested by, among other things, motifs. A coin can be a motif, greed is a theme. [p. 237]

\noindent Theme: A subject (issue, question) that recurs in a narrative through implicit or explicit reference. With motif, theme is one of the two commonest forms of narrarive repetition. Where motifs tend to be concrete, themes are abstract. [p. 242] 
\end{quote}

\citet{abbott2021cambridge} illustrated his definitions of motifs and themes with examples, writing:

\begin{quote}
Windows, for example, are a motif in Wuthering Heights and, given the way Brontë deploys them, they support a highly complex interplay of three themes: escape, exclusion, and imprisonment. When, for another example, the character Barkis in David Copperfield continues to repeat his cryptic phrase, “Barkis is willin’,” it becomes a motif, a signature phrase for the theme of shy, honest-hearted devotion in love that Barkis exhibits in his pursuit of Peggotty. [p. 95]
\end{quote}


Folklorists began creating motif indices to track the presence of specific motifs in specific stories in the late 1800s. The most notable of these motif indices, still in use today, is the Thompson Motif-Index of Folk-Literature \citep[TMI;][]{thompson1977folktale}, in six volumes. Thompson classified motifs into 23 themes, each assigned an alphabetical letter (e.g., A: Mythology, B: Animals, C: Tabu, D: Magic, etc. \citep{thompson1955motif}. Thompson’s system also separates motifs into three types: events, characters, or props. The TMI references tales from over 614 collections, indexed to 46,248 motifs and sub-motifs. 


While the TMI is a seminal work, many folklorists have not found the TMI classification schema as useful for certain cultures. This has been attributed to the TMI not being sensitive to local and regional traditions relevant to the classification of motifs \citep{gay1998review}. In particular, the TMI is more focused on the narrative elements themselves. In contrast, \cite{el1995folk} proposed an approach which, while aligned with TMI subcategories, extended the TMI’s scope to categorize motifs based on their cultural, social, and psychological contextual significance which was not treated by Thompson. For example, a motif can represent a cultural theme, such as \motif{C240}{Tabu: eating with left hand.}~\citep{ElShamy:06}, reflecting that in Islamic-Arabic culture, it is taboo to eat with your left hand. This was expressed in~\citet{irwin2010arabian} as follows:

\begin{quote}
    I went and prepared the necessary food, drink, and so on, which I then presented to him, inviting him to eat in the Name of God. He went to the table and stretched out his left hand, after which he ate with me. This surprised me, and when I had finished, I washed his hand and gave him something to dry it with. I then sat down to talk, after I had offered him some sweetmeats. ‘Sir,’ I said to him, ‘you would relieve me of a worry were you to tell me why you ate with your left hand. Is there perhaps something in your other hand that causes you pain? (Volume 1, Night 25)
\end{quote}

Motifs may also have social significance: \motif{P2.2.1.2}{No low rank person would be sitting down while addressing high rank.} is an example of how motifs can represent the social theme in the narrative~\citep{ElShamy:06}. It is expressed in Night 620 in~\citet{irwin2010arabian} as follows: 

\begin{quote}
    Uthman was both stupid and conceited, and when he arrived at Judar’s palace he saw a eunuch seated on a chair in front of the door. This man did not get up on his arrival, and in spite of the fact that there were fifty men with ‘Uthman, it was as though no one had come. ‘Uthman went up to him and said: ‘Slave, where is your master?’ ‘In the house,’ the eunuch replied, and as he spoke he continued to lounge on his chair. (Volume 2, Night 620)
\end{quote} 

The motif clearly shows how a person's job can shape how society members treat each other. The other motifs in the same category, for instance: \motif{P20.6}{Government by women.} and \motif{P173.9.1}{ Slave purchased.} can be used to trace societal characteristics mentioned in the narrative in a certain period of history about certain communities.

Finally, \motif{U87.1.3}{Clothes make the man.} is an example of how motifs can represent a psychological theme in the narrative in~\citet{irwin2010arabian} as follows:

\begin{quote}
    The weaver went along and saw magnificently dressed people receiving fine foods and being treated with respect by the host because of their splendid clothes. He said to himself: ‘Were I to change my trade for one that would be of less trouble, more prestigious and more rewarding, I could collect a lot of money and buy clothes like these. I would then become important; people would respect me and I would be like these other's. (Volume 1, Night 152)
\end{quote} 

The motif shows how the weaver and other people value each other based on external looks.


While the TMI is the most prominent example of a motif index, there are in fact hundreds of other motif indices that target specific cultures, regions, and language groups. Examples include “A Motif-Index of Traditional Polynesian Narratives” \citep{kirtley2019motif}, “A Type and Motif Index of Japanese Folk-Literature” \citep{ikeda1956type}, the “Motif-index of of Early Irish Literature” \citep{cross1952motif},  and the “Motif-Index of Talmudic-Midrashic Literature” \citep{noy1982motif}, among many others. While the TMI and other motif indices are incredibly useful for the study of motifs, the many different collections that they index vary widely in availability. This is a significant problem for computational work, where access to digitized data is critical to the development, training, and testing of computational systems. This is one reason why no one has previously attempted to develop an automatic motif indexing system.





\subsection{Computational Approaches to Motifs}

\cite {daranyi2010examples} was among the first to call for research into the automatic extraction and annotation of motifs in folklore. In his work he discussed the difficulty of the problem, its potential cross-domain applications (in particular to understanding scientific language), and explored the best tools that can be used to automate the extraction of motifs. He noted that an ability to recognize motifs could underpin a more flexible type of pattern-based search, making it easier to locate specific information within large collections of documents. He and his collaborators also made progress towards using motifs as sequences of “narrative DNA” \citep{daranyi2012detecting,daranyi2012toward}. 

Declerck and collaborators also did computational work on motifs, in particular, converting electronic representations of the TMI to a format that enables multilingual, content-level indexing of folktale texts \citep{declerck2011linguistic,declerck2012multilingual}. This work focuses on the descriptions of motifs and tale types, without reference to the stories.



\cite{yarlott2016learning} planned out a general framework for a potential system for motif extraction and detection. This line of research culminated in a system for detecting usages of motifs in modern text. This work included large-scale annotation showing that human annotators can reliably agree on annotating potential motif phrases. This work produced over 21,000 annotations of motif expressions in modern text to drive the development, training, and testing of the detection system \citep{yarlott2024golem}. Other important contributions were the creation of a pipeline for motif detection and an assessment of its effectiveness across several motif classes \citep[e.g. generalizability:][]{yarlott2022finding,yarlottto2022communicating}. \cite{yarlott2024golem} classified motifs based on the type of usage within a text into four classes: Motific, Referential, Eponymic, or Unrelated. \textsc{motific} refers to a usage which leverages the cultural associations of a motif. \textsc{eponymic} refers to the motif in a name, where the motif is being referenced but the metaphoric meaning is not being used. \textsc{referential} refers to an actual reference to the motif itself, while \textsc{unrelated} refers to a usage unrelated to the cultural group or cannot be established as directly related. 

A related follow-on system was the “Motif Implicit Meaning Extractor (MIME)”, to enable out-of-group persons to better understand and recognize motifs used in text \citep{acharya2024discovering}. The system uses several sources for information on motif meaning, including Wikipedia pages describing the motifs, explicit explanations of motifs from in-group informants, and news/social media posts where the motif is used. Then, it generates a structured report that is simple to read and useful for helping people from other cultures to understand motifs. 


\section{Text \& Index}
\label{sec:data}

As mentioned, one of the major barriers to developing computational motif systems is a lack of easily accessible data, which for the indexing task means a digitized collection of texts that has been indexed by a folklorist. We addressed this problem by selecting The Arabian Nights, which is available in several digitized historical and modern versions, and was indexed by El-Shamy in 2006. This index paired with the text of the Arabian Nights neatly avoid the problem of finding the actual text of widely dispersed tales that vary in availability and digitization. 

\subsection{A Motif Index of The Thousand and One Nights}

A Motif Index of The Thousand and One Nights \citep{ElShamy:06} contains around 5,000 motifs extracted by El-Shamy from the 207 stories of the The Arabian Nights. These motifs overlap with motifs found in the Thompson Motif Index (TMI), while adding new motifs specific to the Arabic content of the stories. El-Shamy generally followed the TMI classification scheme (i.e., 23 themes, each identified by a letter), but he also added categories and information corresponding to dimensions of cultural, social, and psychological contextual significance.

\begin{figure*}[t]
    \centering
    \includegraphics[width=1.0\linewidth]{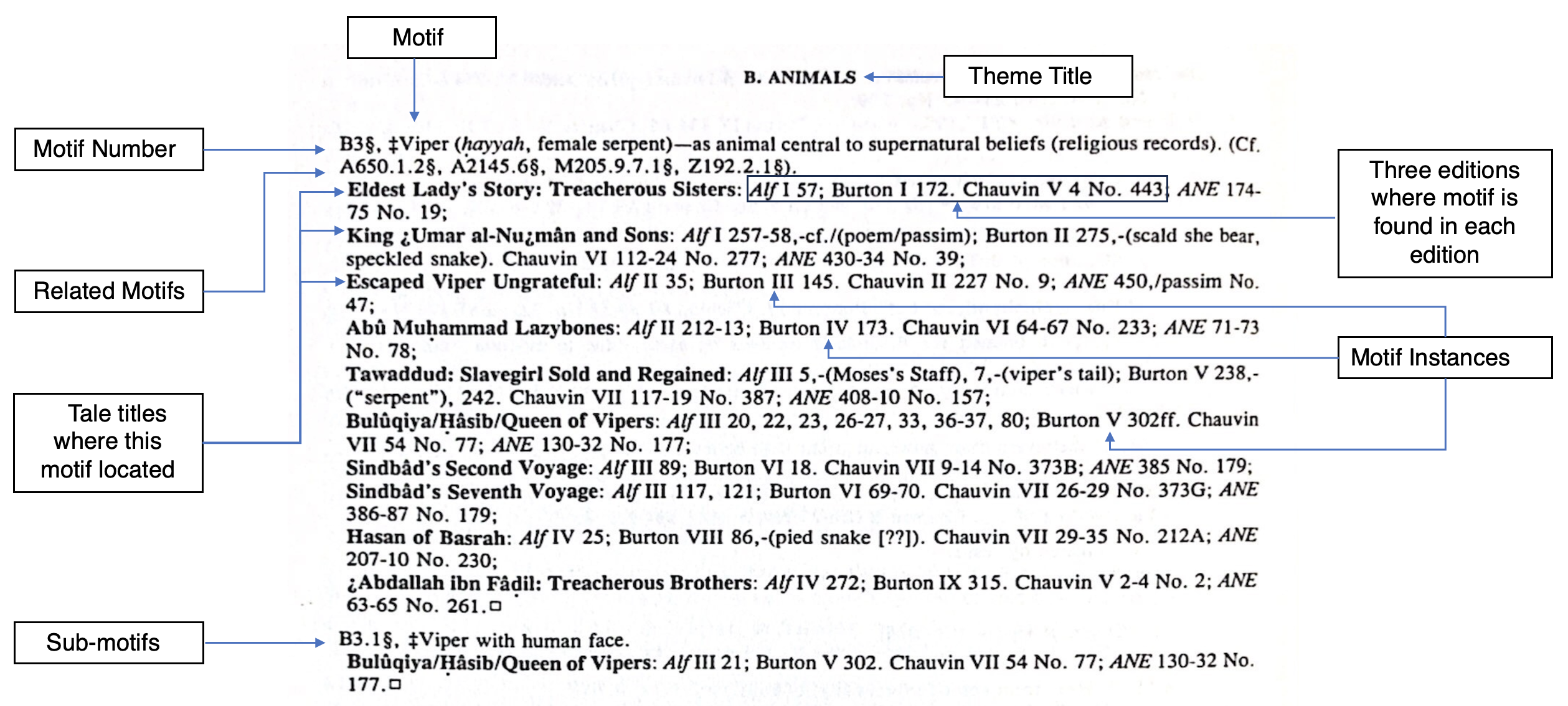}
    \caption{Structure of the entries in \textit{A Motif Index of The Thousand and One Nights}~\citep{ElShamy:06}}
    \label{fig:indexentry}
    \vspace{-1em}
\end{figure*}

The structure of an entry in the motif index is shown in Figure~\ref{fig:indexentry}. The entry for a top-level motif is designated with a number prefixed by the theme letter, such as \motif{B3}{Viper} or \motif{F545}{Other Facial Features}. Each motif entry contains a list of related motifs, a list of tales where the motif is found (“motif expressions”), as well as the page numbers of where that motif expression is found in The Arabian Nights itself or in a reference work. Motifs may also have sub-motifs representing specializations of the main motif, for example, \motif{B3.1}{Viper with human face}, or \motif{F545.1}{Remarkable beard}. Many of the entries contain Arabic words with specialized meanings, such as \textit{hayyah} in the first line of Figure~\ref{fig:indexentry}, which means female serpent.


While El-Shamy’s index is an impressive feat it is important to note that it, like most motif indexes, is not exhaustive, in at least two senses. First, there is no guarantee that El-Shamy identified every single repeated motific pattern in the text. That is, there may be motifs that he did not recognize or document. Moreover, some motifs listed in the index might, when considered by other folklorists, not have been considered “unusual and striking” enough to merit inclusion. Second, El-Shamy’s index, again like most other indexes, only lists a handful of expressions (positive examples) for each motif. For example, for the motif \motif{B3}{Viper} mentioned above, El-Shamy identifies 12 pages, from 12 different tales, where this motif or its sub-motifs appear. However, it can be checked via keyword search and manual confirmation that this motif or its sub-motifs appear in at least 175 places across the text. In each of the tales that are listed, moreover, the motif may appear more than once. Therefore, the motif index lists some number of expressions of each motif, but does not provide an exhaustive list. 

El-Shamy’s index provides page numbers for four different works where each motif is found or discussed. First, he provides the page number of the actual motif expression in the Arabic version of The Arabian Nights \citep{alf_laylah}. Next, he provides page numbers of the motif expressions in Burton’s English translation of \citep{burton1885arabiannights}. Third, he provides a citation to the relevant resumé of the story in Chauvin’s bibliography \citep{Chauvin:92}. Finally, he provides a reference to the relevant resumé for the story in The Arabian Nights Encyclopedia \citep[ANE;][]{marzolph2004arabian}. El-Shamy cites Chauvin and ANE, two encyclopedic books that offer resumés of the stories in French and English, respectively.

\subsection{Burton's Translation of the \textit{The Arabian Nights}}

\begin{figure}[t]
    \centering
\includegraphics[width=\columnwidth]{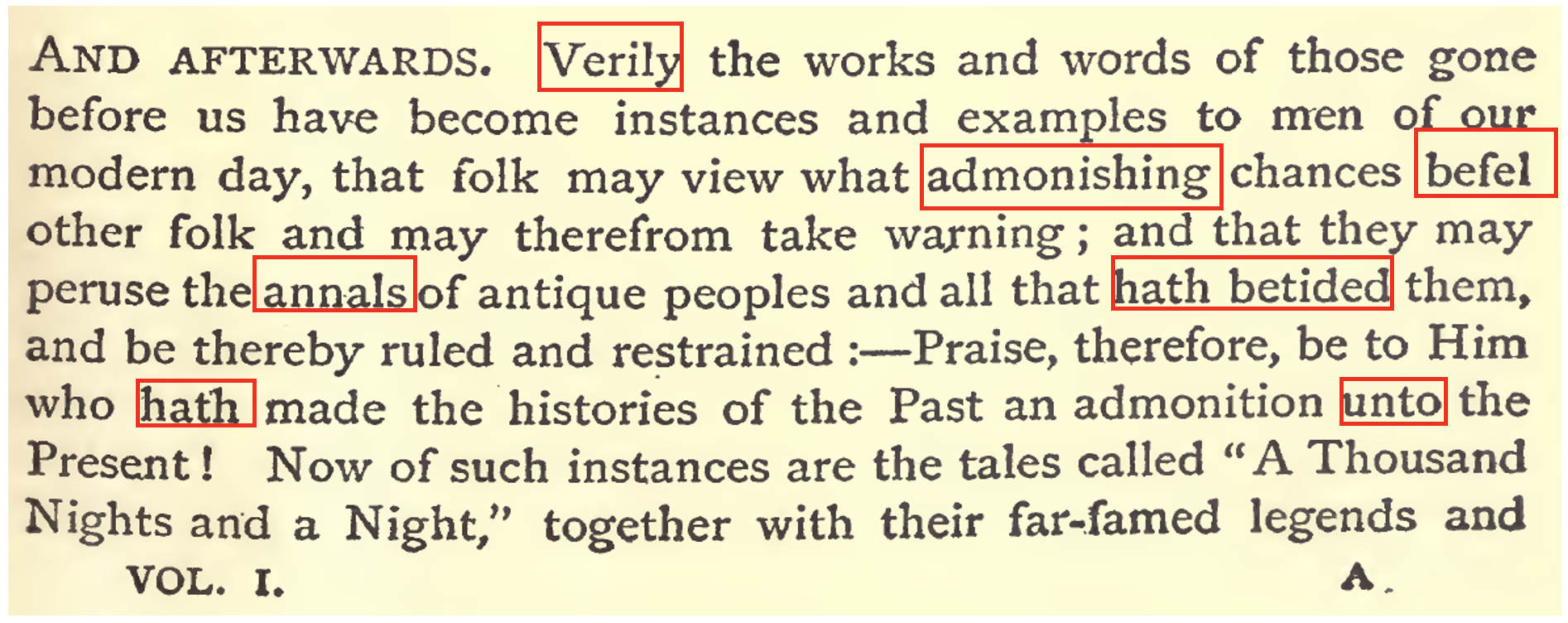}
    \caption{A selection from the available scan of Burton's 1885 translation of \textit{The Arabian Nights}, illustrating the archaic vocabulary (red boxes) and grammar, as well as challenging OCR (note specifically the \textit{f}s in \textit{befel} and \textit{far-famed})}
    \label{fig:burton}
\end{figure}

Burton’s translation consists of 10 volumes plus supplementary volumes and is freely available online. El-Shamy only indexed the 10 volumes, not the supplements, and so we omit the supplements from our analysis. Unfortunately, since Burton’s translation was published in 1885, the text contains relatively archaic English which presents problems for a modern reader and modern NLP systems. Moreover, the available scans of Burton editions are of relatively low quality, resulting in many OCR errors. Figure~\ref{fig:burton} illustrates the archaic language and OCR issues that the text presents. 

To address these problems we found a modern e-book version of The Arabian Nights \citep{irwin2010arabian}. There are three volumes in that edition, comprising a total of 2,706 pages, approximately 45,769 sentences, and 1,398,863 words.

\subsection{Alignment with the Modern Version}

Since El-Shamy’s motif index uses Burton’s translation to identify the location of each motif, we needed to align each page from Burton to its most similar page in the modern edition. For example, motif \motif{B3.1}{Viper with human face} in Figure \ref{fig:indexentry} is identified as appearing in Burton’s volume 5, page 302. Therefore, we needed to find which page in the modern edition aligns with that page. We experimented with two alignment methods: a BERT-based alignment approach and a Needleman-Wunsch algorithm augmented with a synonym detector. 

\subsubsection{BERT-Based Alignment} 

BERT is a transformer-based model that has been used for many NLP tasks. We used BERT to find the best alignment between pages from Burton’s translation to the modern edition by computing the semantic similarity between spans of text. BERT embeds each span into vectors and then we compare vectors using a cosine similarity score~\citep{devlin2018bert}. The goal is to find where each page starts and ends because translations do not have the same length (i.e., page 1 in Burton starts at page 1 in Irwin and ends at page 5). We began by cleaning and lemmatizing the texts, removing all front matter and editorial material, and transforming all text to lowercase. Then we extracted the last 100 words from page 1 of Burton and used a sliding window of 100 words to find the most similar window in Irwin, drawn from the next five pages. We experimented with different numbers of pages, and found 5 to maximize the accuracy. Once the best matching span was found, this was marked as the end of Burton’s page 1 in the Irwin text, and then we repeated the process for the second page, starting from the previously identified page end. We repeated this process until all Burton page start and end spans were placed in alignment with Irwin. We evaluated the accuracy by manually checking 50 pages (5 random pages per volume). This approach achieved 0.93 accuracy. While this accuracy was good, in our opinion it missed too many alignments and degraded the basic quality of the data too much, and so we explored another approach using the Needleman-Wunsch algorithm.

\subsubsection{Needleman-Wunsch Algorithm}

The Needleman-Wunsch Algorithm is a well-known linear programming algorithm which can be used to align sequences of objects \citep{needleman1970general}. For example, it is often used in bioinformatics to align sequences of amino acids. This algorithm takes two sequences and builds a matrix where each cell contains a numerical score, representing a potential match between two items in sequence. A score of 0 means the two items do not match, while 1 indicates the items match perfectly. The algorithm also admits partial matches as well as as an adjustable gap penalty for introducing gaps between the sequences. 

\begin{figure*}[t]
    \centering
    \includegraphics[width=\linewidth]{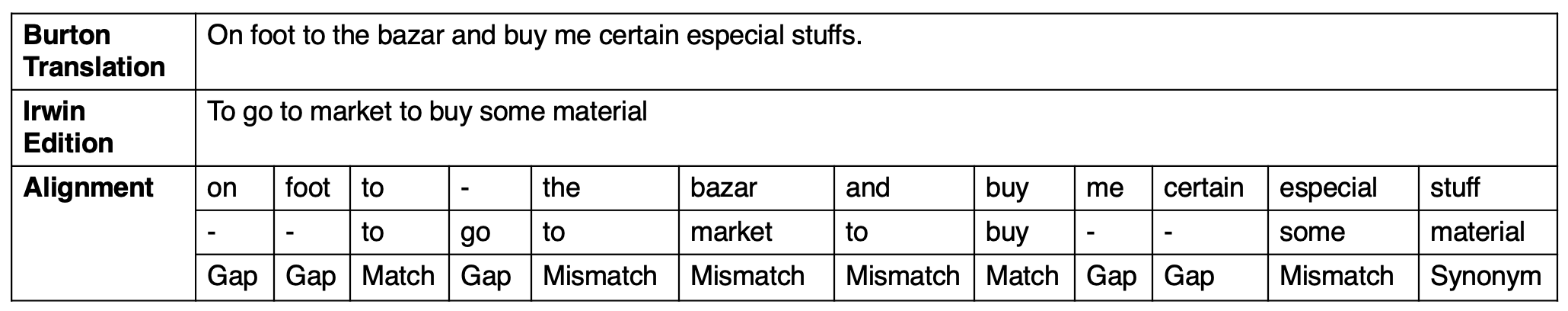}
    \caption{Needleman-Wunsch Example}
    \label{fig:nwexample-v3}
\end{figure*}

We adapted the algorithm to align the word sequences in the two texts, and incorporated a synonym detector to detect when two words were semantically similar (yet not identical) and could therefore be partially aligned. To detect synonyms, we used the Natural Language Toolkit \citep[NLTK;][]{bird2009natural} and WordNet \citep{miller1995wordnet} to lemmatize words to base forms and generate a list of synonyms. When the lemmatized form of two words differed, but the synonym lists of two words intersected, we allowed the two words to placed in alignment with a partial match score of 0.8 (this value was determined experimentally). The time complexity of the Needleman-Wunsch algorithm is $O(mn)$ where $m$ and $n$ represent the lengths of sequences being aligned \citep{needleman1970general}. Because aligning the full sequence of words in both texts would be computationally burdensome (each text is over a million words long), we instead aligned progressive windows of text of 100 words, working off of Burton’s page boundaries, following the same incremental approach as described above for the BERT-based alignment. Figure \ref{fig:nwexample-v3} illustrates the comparison of two texts using the Needleman-Wunsch approach. 

Using the same evaluation procedure as above, our Needleman-Wunsch alignment approach achieved 0.99 accuracy. Therefore, we use the alignment generated by this algorithm as the basis for the remainder of the work.

\section{Analyses of Motif Complexity}
\label{sec:analysis}

In our preliminary investigations of the motif indexing problem, we noted that there were at least two separable dimensions that made the problem difficult: the complexity of the concept expressed by motif, and the complexity of its expression in the text. We analyzed these dimensions in more detail to understand the structure of the problem and guide the collection of the data, so that we had a variety of levels of difficulty and could assess how our techniques fared on the different levels.

\subsection{Conceptual Complexity of Motifs}

The first dimension of difficulty was the conceptual complexity of the motifs. In the motif index, certain motifs expressed quite simple concepts. For example, \motif{B3}{Viper}, mentioned above, refers to just a simple animal type. Other examples, often expressed just as a single word, included \motif{B81}{Mermaid}, \motif{P144.2.1}{Accountant}, or \motif{W217}{Resourcefulness}. However, motifs can be almost arbitrarily complex. An example of a marginally more complex motif is \motif{B3.1}{Viper with a human face}. Examples of highly conceptually complex motifs might be \motif{U10.0.1, What you (deal) to others will be done (dealt) back to you} or \motif{K92.4.1}{Chess game won by distracting opponent’s attention: girl makes seductive gestures (motions) that disorient her male opponent} These latter motifs express highly complex situations involve complex relationships, often with temporal sequencing. 

\begin{figure*}[t]
    \centering
\includegraphics[width=\linewidth]{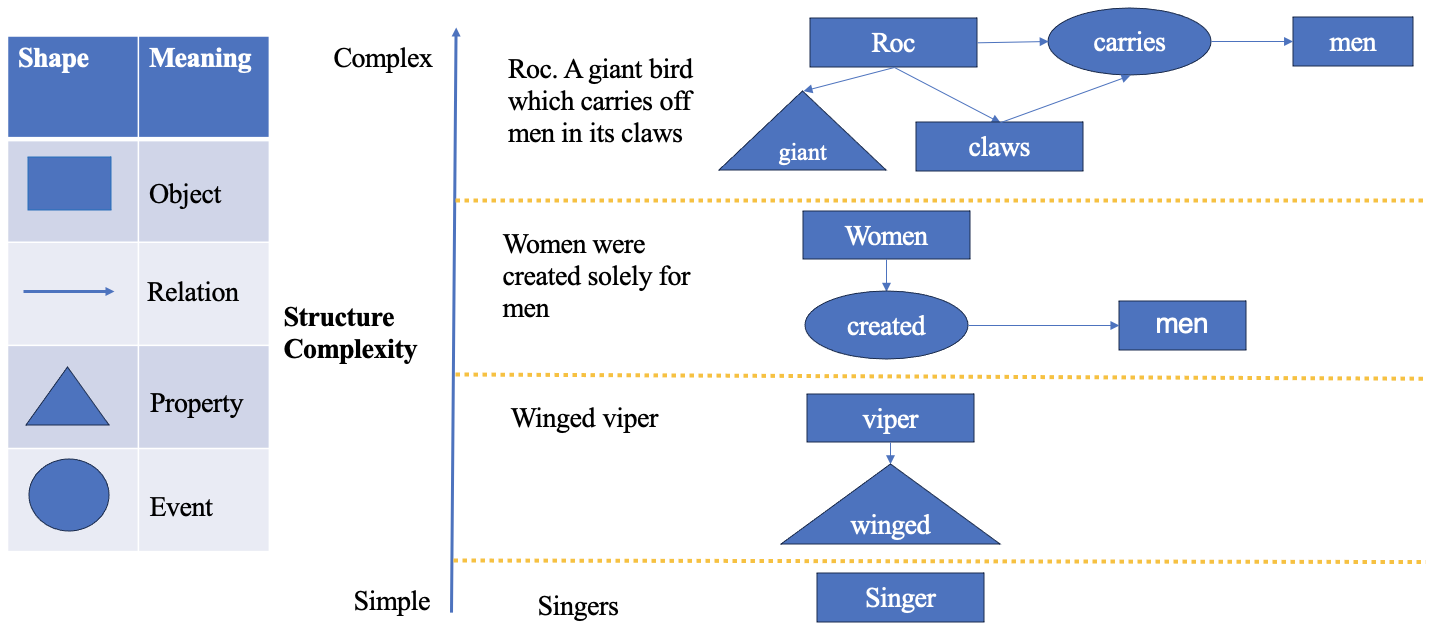}
    \caption{Conceptual Complexity}
    \label{fig:conceptualcomplexity}
\end{figure*}

As shown in Figure~\ref{fig:conceptualcomplexity} we used a semantic-net-style analysis to explore the conceptual complexity of a selection of motifs from the index. Each analysis is a small graph, where nodes might represents objects, properties, or events, and arrows represent relationships between them (such as an object having a property, an object being an agent or patient of an event, or one event temporally following another). Using such an analysis we could array motifs along a number of possible dimensions of conceptual complexity, for example, number of objects, number of properties, number events, length or shape of temporal sequence, number of total nodes in the graph, and so forth. While this analysis will be interesting to explore in future work, here we were interesting in defining the task and setting a baseline for the problem, so we only divided motifs into “simple” and “complex” conceptual complexity. In general, if the conceptual graph had more than two nodes, we considered it as “complex”; otherwise we considered it “simple.”

\subsection{Expression Complexity of Motifs}
The second dimension of complexity for motifs was how they were expressed in the text. A simple motif such as \motif{B3}{Viper} might easily be expressed quite simply in the text using the single word \textit{viper} or \textit{snake}. Simple motifs simply expressed are quite easy to find. However, even conceptually simple motifs might be expressed in quite complex and even obscure ways, as shown in this examples of the motif \motif{W210}{Daydreaming}, which appears over the span of nearly 5 pages in the Irwin text: 

\begin{quote}
This glass represents my capital of a hundred dirhams. I shall sell it for two hundred and then use the two hundred to buy more, which I shall sell for four hundred. I shall go on buying and selling until I have great wealth, and then I shall buy all kinds of goods, jewels and perfumes, and make an enormous profit. After that, I shall buy a fine house, with mamluks, horses and saddles of gold. I shall eat and drink and invite home every singer in the city, whether male or female. My capital, God willing, will come to a hundred thousand dirhams. . . [description of the daydream goes on for five more pages] (Volume 1, Night 32)
\end{quote}

The same potential for complexity of expression holds true for more conceptually complex motifs as well. 

Again, rather than proposing a fine-grained scale of expression complexity (which would be interesting to explore in future work), we started by separating motif expressions into “simple” or “complex”. The rule for this was less hard and fast than for conceptual complexity. In general, if the expression on the page was the same or very close to the description of the motif in the index entry, we labeled it “simple”. Otherwise, we  labeled it ``complex''. We did not rigorously evaluate this separation, as it was merely to guide our evaluation efforts in this first-of-its-kind study. In future work we would like to explore this idea more deeply.

\subsection{Combining the Complexity Dimensions}

Combining the two dimensions of conceptual complexity and expression complexity gives a four-way categorization of motif expressions in the text. We can intuit the difficulty of finding these different types as discussed below.

\subsubsection{Simple Structure, Simple Expression}
This is the easiest kind of motif where both the structure and the way the motif is expressed in the narrative are simple. For example, \motif{P234.0.3.1}{Seven daughters} is simply structured and expressed easily: \textit{The Almighty provided him with seven daughters\ldots}~\citep[Volume~3, Night~784]{irwin2010arabian}.

\subsubsection{Simple Structure, Complex Expression}
Similar to the previous class except that the expression is difficult. In other words, the conceptual structure of the motif is simple (and is expressed in the index in only one or a few words), but in the text the motif is expressed indirectly, obscurely, or across many pages. For example, the motif \motif{W217}{Resourcefulness} is conceptually simple, but one expression runs \textit{The mamluks delightedly agreed that this was a good plan, and there and then they began to cut logs for the raft and to twist ropes to bind them together. They worked on this for a month, taking back firewood for the princess’s kitchen each evening and devoting the rest of the day to the raft.}~\citep[Volume~3, Night~766]{irwin2010arabian}.

\subsubsection{Complex Structure, Simple Expression}
For this type of expression, the conceptual structure is complex, but still the expression of the motif in the text closely follows what is found in the index, and so keyword retrieval suffices to find them. For example, the motif \motif{E68}{Apparently dead persons revived when certain thing happens. Proper prince appears, or the like.} is not simply structured but can be expressed almost exactly as is found in the index, as in \textit{You Muslims, you soldiers, have you ever in your lives seen a man die and then come back to life?}~\citep[Volume~1, Night~34]{irwin2010arabian}.

\subsubsection{Complex Structure, Complex Expression} 
This is the most difficult type of motif expression to find. The motif is conceptually complicated and motif expression is also complex. For example, the motif \motif{U10.0.1}{What you (deal) to others will be done (dealt) back to you} is found in the text as follows:

\begin{quote}   
I took hold of the horse and mounted it. It didn’t move and so I kicked it, and when it still refused to move, I took the whip and struck it. It didn’t move and so I kicked it, and when it still refused to move, I took the whip and struck it. As soon as it felt the blow, it neighed with a sound like rumbling thunder and, opening up a pair of wings, it flew off with me, carrying me up into the sky way above the ground. After a time, it set me down on a flat roof and whisked its tail across my face, striking out my right eye and causing it to slide down my cheek. It then left me and I came down from the roof to find the ten one-eyed youths. No welcome to you, they said. Here I am, I replied. I have become like you, and I want you to give me a tray of grime with which to blacken my face and to let me sit with you. No, by God, they said, you may not do that. Get out!~\citep[Volume~1, Night~16]{irwin2010arabian}.
\end{quote} 

The motif is found in across approximately 190 words and 10 sentences, where none of the words in the motif definition are found directly in the expression.

\section{Annotation}
\label{sec:annotation}

To generate data for validation, training, and testing, we pursued an iterative, machine-in-the-loop annotation strategy, where we used a baseline retrieve-and-rerank system to identify sentences with potential motif expressions, and then manually labeled those to evaluate the performance of the system. This evaluation drove refinements to retrieve-and-rerank and later methods. This process of data exploration and annotation was necessary because, as mentioned previously, El-Shamy's index is not exhaustive (it only lists a few positive examples for each motif), and so while we knew where some positive expressions appeared, we had to search manually for others. This approach allowed us to quickly build up a large set of sentences annotated for the presence or absence of specific motifs, while simultaneously evaluating our systems to drive further improvement. The details of the retrieve-and-rerank system are given in \nameref{sec:retrieve-and-rerank}.

We retrieved and annotated individual sentences from the text. This approach misses any motif expressions that are larger than a sentence in length. We gave several examples of motif expressions longer than a sentence above, in \nameref{sec:analysis}. However, in the end we had only 68 multi-sentence motif expressions in our set, out 2,670 positive examples. Furthermore, of these 68 multi-sentence positive example, 39 of them had a sentence which, considered by itself, communicated the motif well enough to identify the motif being used. This means that, by our estimate, only 1.1\% of motif expressions actually require one to consider information spanning multiple sentences, and so  this phenomenon is relatively rare. Therefore, for this initial study we focused only on identifying motif expressions within the bound of a single sentence. Detecting multi-sentence motif expressions we leave for future work.


\subsection{Motif Selection}

We began by selecting motifs that were represented by a single word in the index to test how well simple keyword retrieval approaches performed. Investigating false positives led us to the conceptual and expression complexity analysis described above (\nameref{sec:analysis}). Following this, we sought to diversify the conceptual complexity of our selection, by selecting additional motifs with multiple words (such as a noun and an adjective), as well as motifs with longer descriptions of 3 or more words. The number of motifs that fell into simple or complex structure is shown in Table~\ref{tab:data}.

\begin{table}[t]
\center
\caption{Annotation Data}
\begin{tabular}{rl}
\toprule
\bf Data & \bf Count \\
\midrule
Sentences in the Text                         & 45,769 \\
Tokens in the Text                            & 1,398,863 \\
\midrule
Unique Motifs                                & 200 \\
Positive Examples in the Index                       & 266 \\
Motifs with 1 Positive Example in the Index            & 175 \\
Motifs with more than 1 Positive Example in the Index           & 26 \\
Motifs, Simple Structure                     & 142 \\
Motifs, Complex Structure                    & 58 \\
\midrule
Annotated Sentences-Motif pairs              & 58,450 \\
Unique pairs       & 26,262 \\
Percentage of Sentences with an annotation & 57\% \\
Tokens in annotated sentences                & 1,412,912 \\
Sentences-Motif pairs, Positive Examples             & 2,670 \\
Positive Examples Total Tokens                       & 88,523\\
Sentences-Motif pairs, Negative Examples             & 55,779 \\
\bottomrule
\end{tabular}
\label{tab:data}
\end{table}

\subsection{Annotator Training}
Annotation was performed by four annotators, the first of which was the first author of the paper, and the remaining three were paid research assistants (one MS student and two undergraduates). Development of the annotation guide was done in an incremental fashion, by first starting from positive examples identified by El-Shamy to learn how motifs appeared in the text, and reasoning by analogy to new sentences identified by versions of the system. In early stages the annotators worked as a team to go through a small sample annotation and compare notes, capturing their observations in the annotation guide. Every week the annotators each reviewed approximately 1,500 sentence-motif pairs that had been retrieved by the latest version of the system. Approximately half of these pairs were double-annotated, so every week the team produced approximately 2,250 annotations of sentence-motif pairs. The team also held weekly half-hour adjudication sessions to review disagreements. These sessions were also used to refine the annotation guide. The annotation guidelines are given in Appendix \ref{sec:annotationguidelines}.

\begin{table}[t]
\centering
\caption{Numbers of Motif by Motif Expression Complexity}
\begin{tabular}{c|c@{~~~~~~~~}c@{~~~~~~~~}c|c}
    \toprule
                   & \multicolumn{4}{c}{\bf Expression Complexity} \\
    \bf Conceptual Complexity & \multicolumn{4}{c}{} \\ \cmidrule(lr){2-5}
    \bf $\downarrow$ & \bf Simple & \bf Simple \& & \bf Complex & \\
                     & \bf Only   & \bf Complex   & \bf Only    & \bf Total \\
    \midrule
    \bf Simple     & 34          & 73               & 35      & 142\\
    \bf Complex    & 11          & 25               & 22      & 58 \\
    \midrule
    \bf Overall    & 45          & 98               & 57      & 200 \\
    \bottomrule
\end{tabular}
\label{tab:motifsbycomplexity}
\end{table}
\subsection{Summary of the Annotated Data}

Summaries of the annotated data are shown in Tables~\ref{tab:data}--\ref{tab:motifexpressions}. In particular. Table~\ref{tab:data} shows the overall counts of number of motifs and number of sentences annotated, with both positive and negative examples. Table~\ref{tab:motifsbycomplexity} breaks down motifs by which ones are found only simply expressed in our data, which are only expressed in complex ways, and which are found in both ways. Table~\ref{tab:motifexpressions} breaks down the identified motif expressions (all positive examples) by conceptual complexity of the expressed motif and expression complexity. 

\begin{table}[t]
\begin{center}
\caption{Total Motif Expressions in Each Category}
\begin{tabular}{c|cc|c}
    \toprule
    \bf Conceptual Complexity & \multicolumn{2}{c|}{\bf Expression Complexity} \\
    & \multicolumn{2}{c|}{} \\
    $\downarrow$ & \bf Simple & \bf Complex & \bf Overall \\
    \midrule
    \bf Simple     & 1,467       & 689      & 2,156 \\
    \bf Complex    & 211       & 303      & 514 \\
    \midrule
    \bf Overall    & 1,678       & 992      & 2,670  \\
    \bottomrule
\end{tabular}
\label{tab:motifexpressions}
\end{center}
\end{table}

\subsection{Inter-Annotator Agreement}

We calculated the inter-annotator agreement (also known as the inter-rater reliability) using 5,548 sentences and 60 motifs. Overall we have reached Cohen's Kappa scores for each category shown in Table~\ref{tab:kappascore}. As would be expected, agreement was lowest for motifs that have been expressed in a complex manner, with overall agreement of 0.53. Nevertheless, this still represents ``moderate'' agreement~\citep{landis1977measurement}. Agreement for simply expressed motifs with simple conceptual complexity is 0.87 (``perfect'' agreement) and that for simply expressed motifs with complex conceptual complexity is 0.73 (``substantial'' agreement). Overall agreement across the set is 0.72, which indicates ``substantial'' agreement. These numbers indicate that the task can be done with a high degree of repeatability across annotators. 

\begin{table}[h]
\centering
\caption{Cohen's Kappa Score for Each Category}
\begin{tabular}{c|c@{~}c|c@{~~~~}c}
    \toprule
    \bf Conceptual Complexity & \multicolumn{2}{c|}{\bf Expression Complexity} \\
                              & \multicolumn{2}{c|}{}  & & \bf Total \\
    $\downarrow$ & \bf Simple & \bf Complex  & \bf Overall & \bf Motifs \\
    \midrule
    \bf Simple       & 0.87     & 0.51         &   0.71      & 50 \\
    \bf Complex      & 0.73     & 0.54         &   0.72      & 10 \\
    \midrule
    \bf Overall      &  0.85    & 0.53         &   0.72      & -  \\
    \bf Total Motifs & 30       & 30           & -           & 60 \\
    \bottomrule
\end{tabular}
\label{tab:kappascore}
\end{table}





\section{Methods}
\label{sec:methods}

We evaluated five approaches to solving the task. We began with a retrieve-and-rerank approach (\nameref{sec:retrieve-and-rerank}). Then we developed embedding comparison techniques using both off-the-shelf embedding models (\nameref{sec:offtheshelf}) and fine-tuned embedding models (\nameref{sec:finetuned}). Finally, generative approaches using off-the-shelf LLMs (\nameref{sec:generative})

\subsection{Retrieve-and-Rerank Implementation}
\label{sec:retrieve-and-rerank}
We initially explored a retrieve-and-rerank approach to detect motif expressions in the text. This technique has been used in a variety of Information Retrieval (IR) and Question Answering tasks~\citep{sbert_retrieve_rerank}. In retrieve-and-rerank, one first does a high recall retrieval (possibly in several stages), and then reranks the returned candidates to push better matches to the top of the queue. Performance of such an approach can be measured by a recall measured versus rank in the final list. 

We initially chose retrieve-and-rerank for two reasons. First, a keyword-based retrieval approach is a natural first step when one has no annotated data, is computationally inexpensive, and can serve as the first stage of a retrieve-and-rerank approach. Second, keyword-based retrieval is a natural complement to off-the-shelf semantic similarity scorers, which also can be found in computational inexpensive but high performing versions. Both of these high-recall approaches can then be easily combined with an off-the-shelf small reranker. This system could be profitably combined with our annotation process to produce quick turnaround between annotation stages, allow us to quickly gather a large body of annotated data for use in developing more high performing approaches.

Thus our retrieve-and-rerank approach used two recall stages (lexical and semantic) followed by a reranking stage. This two-pronged approach also naturally allowed us to explore how performance varied with motifs with different conceptual and expression complexities.




\subsubsection{Lexical Retrieval}

In this stage, we searched for the target motifs based on lexical search, similar to that used in common text search engines, e.g., Lucene~\citep{mccandless2010lucene}. In particular, we employed the Okapi Best Matching algorithm~\cite[BM25;][]{robertson2009probabilistic}, which is a standard ranking method that search engines employ to determine how relevant pages are to a certain search query. BM25 uses an equation that takes into consideration term frequency (TF), inverse document frequency (IDF), and document length. We limited the output of this stage to 100 possible sentences, which are evaluated in the reranking step. Note that lexical retrieval takes into account actual lexical items found in the motif definition, not synonyms or rewordings. Table~\ref{tab:lexsemscores} shows the recall score for each category.

\subsubsection{Semantic Retrieval} 

In contrast to lexical retrieval, with semantic retrieval in this stage we sought sentences which lay close to the motif definition in an embedding space. We individually embedded the motif definition and all sentences in the text, using the \texttt{all-mpnet-base-v2} model, which is one of the highest performing neural sentence embedding model that is a relatively small model~\citep{reimers2019sentence}. As in the lexical stage, we then retrieved 100 sentences which were closest to the motif definition as measured by cosine similarity~\citep{rahutomo2012semantic}.  Table~\ref{tab:lexsemscores} shows the recall score for each category.

\begin{table}
\centering
\caption{Recall of each retrieval stage (lexical and semantic) for each category of motif expression}
\begin{tabular}{c@{~}c|cc|c}
\toprule
\bf Conceptual Complexity &      & \multicolumn{2}{c|}{\bf Expression Complexity} \\
                          &      & \multicolumn{2}{c|}{}  \\
$\downarrow$   &      & \bf Simple & \bf Complex & \bf Overall \\
\midrule
\multirow[c]{3}*{\bf Simple}     
               & Lex. & 0.48 & 0.16 & 0.38 \\
               & Sem. & 0.69 & 0.60 & 0.66 \\
               & Both & 0.82 & 0.40 & 0.69 \\
\midrule
\multirow[c]{3}*{\bf Complex}
               & Lex. & 0.61 & 0.12 & 0.32 \\ 
               & Sem. & 0.58 & 0.59 & 0.59 \\
               & Both & 0.73 & 0.21 & 0.42 \\
\midrule
\multirow[c]{3}*{\bf Overall}
               & Lex. & 0.50 & 0.15 & 0.37 \\
               & Sem. & 0.68 & 0.60 & 0.65 \\
               & Both & 0.81 & 0.34 & 0.64 \\
\bottomrule
\end{tabular}
\label{tab:lexsemscores}
\end{table}

\subsubsection{Reranking}

Together the lexical and semantic retrievals output 100 sentences each. The combined recall of the two stages is shown in Table~\ref{tab:lexsemscores}. We chose this number because most of the motifs we examined have many fewer than 200 positive examples across the whole text. This number, then, is a kind of hyper parameter that depends on that data, and could easily be increased or decreased based on data size, motif characteristics, or other factors. To rerank the 200 item list, we used BERT in a cross-encoder setup~\citep{devlin2018bert}. A cross-encoder approach is especially useful for assessing the similarity of a pair of texts; in our case we cross-encoded the motif description with each sentence to be evaluated. To compute the cross-encoder result each sentence was concatenated to the motif description, and fed to BERT to produce a yes/no judgment. Table~\ref{tab:r-and-r} shows the $F_1$ score for each category after the reranking stage. 

\begin{table}[t]
\centering
\caption{Overall System Evaluations: $F_1$ (precision / recall)}
\begin{tabular}{c|cc|c}
\toprule
\bf Conceptual  & \multicolumn{2}{c|}{\bf Expression Complexity} \\
\bf Complexity                  & \multicolumn{2}{c|}{} \\
$\downarrow$              & \bf Simple     & \bf Complex  & \bf Overall\\
\midrule
\multirow[c]{1}*{\bf Simple} & \rpf{94}{25}{39}  & \rpf{62}{20}{30}  & \rpf{78}{23}{35} \\
\midrule
\multirow[c]{1}*{\bf Complex} & \rpf{51}{29}{37}  & \rpf{50}{29}{37}  & \rpf{50}{29}{37} \\
\midrule
\multirow[c]{1}*{\bf Overall} & \rpf{72}{27}{38}  & \rpf{56}{25}{34}  & \rpf{64}{26}{36} \\
\bottomrule
\end{tabular}
\label{tab:r-and-r}
\end{table}

\begin{table}[t]
\centering
\caption{Breakdown into training / validation / testing}
\begin{tabular}{@{}c|c@{~~}c|c@{}}
    \toprule
    \bf Conceptual Complexity & \multicolumn{2}{c|}{\bf Expression Complexity} \\
                              & \multicolumn{2}{c|}{} \\
    $\downarrow$ & \bf Simple     & \bf Complex  & \bf Overall \\
    \midrule
    \bf Simple     & \sss{22}{3}{9} & \sss{81}{17}{10} & \sss{103}{20}{19} \\
    \bf Complex    & \sss{6}{3}{2}   & \sss{31}{7}{9} & \sss{37}{10}{11} \\
    \midrule
    \bf Overall    & \sss{28}{6}{11} & \sss{112}{24}{19} & \sss{140}{30}{30} \\
    \bottomrule
\end{tabular}
\label{tab:databreakdown}
\end{table}





\subsection{Embedding Models}
Given the difficulty of the task for the retrieve-and-rerank approach, especially in the case of lexical retrieval, we also experimented with a pure embedding based approaches. We first experimented with off-the-shelf embedding models, then fine-tuned embedding models. 


\subsubsection{Off-the-Shelf Embedding Models}
\label{sec:offtheshelf}


We experimented with a pure embedding based approaches. We first experimented with off-the-shelf embedding models, specifically Mistral \cite[\texttt{mistral-embed};][]{mistral_embeddings}, Gemini~\cite[\texttt{text-embedding-004};][]{text-embedding-004}, Nvidia's NV-Embed~\cite[\texttt{NV-Embed-v2};][]{nv-embed-v2}, Jina AI~\cite[\texttt{jina-embeddings-v3};][]{sturua2024jina}, and T5~\cite[\texttt{sentence-t5-
base}; ][]{ni2021sentence}. Despite reasonable performance, the task is still challenging, with performance maxing out at 0.65 $F_1$. We used the models (either the API or a locally installed model, if available) to generate text embeddings for each sentence, and used the cosine similarity scores combined with a threshold to label sentences as positive or negative. We identified the threshold by first averaging the cosine similarities of all the positive examples (across all motifs), then averaging the cosine similarities of the negative examples, and then finding the midpoint. Using this procedure, we determined a threshold of 0.73 for Mistral model, 0.46 for Gemini, 0.25 for Nvidia, 0.32 for Jina, 0.77 for T5. Results for the embedding models are shown in Table~\ref{tab:results}.

Since our annotated sentence-level data is imbalanced, with a majority of negative examples (i.e., 55,779 sentences with no motif, compared with 2,670 sentences with a motif expression), we resampled the negative examples, resulting in 5,340 total examples. In this set, each motif has an equal number of positive and negative sentences. We split the resampled dataset into training, validation, and test sets (shown in Table~\ref{tab:databreakdown}) based on our two dimensions of the complexity of expressions and structures. Each training, testing, or validation comprises a unique list of motifs so that each model is tested on an unseen list of motifs. 

\subsubsection{Fine-Tuned Embedding Models}
\label{sec:finetuned}

We also experimented with fine-tuning embedding models. As a baseline, we first fine-tuned SBERT~\citep{reimers2019sentence}, which is pre-trained on a large corpus of text data for semantic similarity tasks between sentences. The off-the-shelf Jina AI model performed quite well, and we would have liked to fine-tune that model, but it was not made available to us. So we fine-tuned the open T5 model, which has a similar architecture to Jina. We used the cosine similarity loss function with a batch size of 16, and trained for 5 epochs for fine-tuning, which is appropriate in cases where a dataset consists of pairs of text and a label~\citep{sbert_retrieve_rerank}. 

We used the fine-tuned model to generate embeddings for the testing set. Then, cosine similarity was used to measure the similarity of the generated vectors from the fine-tuned model. Following the same threshold procedure described previously, we determined a threshold of 0.32 for fine-tuned SBERT and 0.45 for fine-tuned T5. Results of these experiments are shown in Table~\ref{tab:results}. 

\begin{table*}[!ht]
\centering
\caption{Overall System Evaluations: $F_1$ (presion / recall)}
\label{tab:results}
\begin{tabular}{c@{~}r|ccc|ccc}
\toprule
\multicolumn{2}{l|}{\bf Conceptual Complexity} & \multicolumn{3}{c|}{\bf Expression Complexity} & \multicolumn{3}{c}{\bf Model Type} \\
\multicolumn{2}{l|}{} & \multicolumn{3}{c|}{} &\multicolumn{3}{c}{} \\
$\downarrow$ & {} & \bf Simple & \bf Complex & \bf Overall    \\
\midrule
\multirow[c]{8}*{\vspace{-6em}\rotatebox{90}{\bf Simple}} 
& text-embedding-4     & \rpf{58}{71}{64}  & \rpf{63}{49}{55}  & \rpf{60}{60}{60} & Embedding\\
& NV-Embed-v2          & \rpf{54}{67}{60}  & \rpf{66}{47}{55}  & \rpf{60}{57}{58} & Embedding\\
& mistral-embed        & \rpf{81}{44}{58}  & \rpf{78}{21}{34}  & \rpf{80}{33}{47} & Embedding\\
& jina-embeddings-v3   & \rpf{74}{71}{72}  & \rpf{73}{46}{57}  & \rpf{74}{59}{65} & Embedding\\
& sentence-t5-base     & \rpf{83}{66}{73}  & \rpf{89}{42}{57}  & \rpf{86}{54}{66} & Embedding\\
& SBERT-FT             & \rpf{72}{40}{51}  & \rpf{66}{17}{28}  & \rpf{69}{29}{40} & Fine-Tuned\\
& sentence-t5-base-FT  & \rpf{89}{72}{80} & \rpf{72}{43}{54}  & \rpf{81}{57}{67}  & Fine-Tuned\\
& Mistral-Zero-Shot        & \rpf{56}{93}{70}  & \rpf{20}{66}{31}  & \rpf{47}{89}{61} & Zero-Shot\\
& Llama3-Zero-Shot        & \rpf{81}{79}{80}  & \rpf{73}{57}{64}  & \rpf{79}{72}{75} & Zero-Shot\\
& Mistral-Few-Shots     & \rpf{81}{80}{81}  & \rpfb{84}{71}{77}  & \rpf{82}{78}{80} & 2-Shots\\
& Llama3-Few-Shots    & \rpf{94}{71}{81}  & \rpf{96}{55}{70}  & \rpf{96}{55}{70} &  2-Shots\\
& Mistral-FT            & \rpf{83}{90}{86}  & \rpf{58}{74}{65}  & \rpf{76}{86}{81} &  Fine-Tuned\\
& Llama3-FT            & \rpfb{91}{89}{90}  & \rpf{67}{81}{73}  & \rpfb{87}{87}{87} & Fine-Tuned\\

\midrule
\multirow[c]{8}*{\vspace{-6em}\rotatebox{90}{\bf Complex}} 
& text-embedding-004   & \rpf{83}{58}{68}  & \rpf{69}{49}{57}  & \rpf{76}{53}{63}  & Embedding\\
& NV-Embed-v2          & \rpf{80}{57}{67}  & \rpf{74}{51}{60}  & \rpf{77}{54}{64}  & Embedding\\
& mistral-embed        & \rpf{71}{33}{45}  & \rpf{60}{26}{37}  & \rpf{65}{30}{41}  & Embedding\\
& jina-embeddings-v3   & \rpf{71}{55}{62}  & \rpf{83}{55}{66} & \rpf{77}{55}{64}  & Embedding\\
& sentence-t5-base     & \rpf{83}{46}{59}  & \rpf{86}{43}{57}  & \rpf{84}{44}{58}  & Embedding\\
& SBERT-FT             & \rpf{68}{34}{46}  & \rpf{80}{34}{48}  & \rpf{74}{34}{47} & Fine-Tuned\\
& sentence-t5-base-FT  & \rpf{85}{59}{70} & \rpf{66}{49}{56}  & \rpf{76}{54}{63} & Fine-Tuned\\
& Mistral-Zero-Shots        & \rpff{53}{1}{69}  & \rpf{31}{91}{46}  & \rpf{43}{97}{60} & Zero-Shott\\
& Llam3-Zero-Shot        & \rpff{53}{1}{69}  & \rpf{51}{85}{64}  & \rpf{52}{93}{67} & Zero-Shot\\
& Mistral-Few-Shots        & \rpf{68}{93}{78}  & \rpf{45}{84}{59}  & \rpf{57}{89}{70} & 2-Shots\\
& Llama3-Few-Shots        & \rpf{82}{80}{81}  & \rpf{62}{70}{66}  & \rpf{73}{76}{75} & 2-Shots\\
& Mistral-FT            & \rpfb{80}{94}{86}  & \rpfb{60}{91}{72}  & \rpfb{71}{93}{80} & Fine-Tuned\\
& Llama3-FT            & \rpf{84}{86}{85}  & \rpf{59}{86}{70}  & \rpf{73}{86}{79}  & Fine-Tuned\\

\midrule
\multirow[c]{8}*{\vspace{-6em}\rotatebox{90}{\bf Overall}} 
& text-embedding-004   & \rpf{70}{64}{66}  & \rpf{66}{49}{56}  & \rpf{68}{57}{61} & Embedding\\
& NV-Embed-v2          & \rpf{67}{62}{63}  & \rpf{70}{49}{58}  & \rpf{69}{55}{61} & Embedding\\
& mistral-embed        & \rpf{76}{39}{51}  & \rpf{69}{24}{35}  & \rpf{73}{31}{44} & Embedding\\
& jina-embeddings-v3   & \rpf{72}{63}{67}  & \rpf{78}{51}{61}  & \rpf{75}{57}{65} & Embedding\\
& sentence-t5-base     & \rpf{83}{56}{66}  & \rpf{87}{42}{57}  & \rpf{85}{49}{62} & Embedding\\
& SBERT-FT             & \rpf{70}{37}{48}  & \rpf{73}{26}{38}  & \rpf{71}{31}{44} & Fine-Tuned\\
& sentence-t5-base-FT  & \rpf{87}{66}{75} & \rpf{69}{46}{55}  & \rpf{78}{56}{65} & Fine-Tuned\\
& Mistral-Zero-Shot        & \rpf{56}{94}{70}  & \rpf{23}{75}{36}  & \rpf{46}{90}{61}& Zero-Shot\\
& Llama3-Zero-Shot        & \rpf{77}{81}{79}  & \rpf{66}{62}{64}  & \rpf{74}{74}{74} & Zero-Shot\\
& Mistral-Few-Shots        & \rpf{79}{82}{81}  & \rpfb{72}{73}{73}  & \rpf{77}{79}{78} & 2-Shots\\
& Llama3-Few-Shots        & \rpf{92}{72}{81}  & \rpf{85}{58}{69}  & \rpf{90}{67}{77} & 2-Shots\\
& Mistral-FT            & \rpf{82}{91}{86}  & \rpf{58}{78}{67}  & \rpf{75}{88}{81}& Fine-Tuned\\
& Llama3-FT            & \rpfb{91}{89}{90}  & \rpf{64}{82}{72}  & \rpfb{84}{86}{85} & Fine-Tuned\\
\bottomrule
\end{tabular}
\end{table*}

\subsection{Generative Approaches}
\label{sec:generative}

Finally, we explored generative approaches, which involve prompting an off-the-shelf model using Zero- and Few-shot paradigms~\citep{mann2020language} and fine-tuning large language models (LLMs) to solve the task. We experimented with open-source LLMs using Zero-shot, Few-shot, and fine-tuning on two large language models (LLMs) using Low-Rank Adaptation (LoRA): Mistral \citep[Mistral-7B-Instruct-v0.3; May 2024:][]{mistral7b_instruct_v0_3} and Llama \citep[Llama-3.1-8B-Instruct; July 2024:]{meta2024llama31_8b_instruct}. We used the same training/validation/test dataset we used previously with the fine-tuned embedding models shown in Table~\ref{tab:databreakdown}.  

\subsubsection{Zero-shot}
First we experimented with the two LLMs using a Zero-shot approach by inserting the motif and a sentence in the prompt. The goal was to see how the baseline LLMs classify whether a sentence contains a motif or not, giving positive and negative sentences separately without the models' past knowledge of the motif in question. We used and forced the models to output a single token (Yes or No) using the following prompt:

\begin{quote}
\tt Task: Decide if the motif is present in the \\
sentence. Rules: Answer ONLY ``Yes'' or ``No''. \\ 
Do not explain.
\end{quote}

To avoid randomness, we set the temperature to 0 so the model produces the same output for the same input. We also restrict the model to generate only one token (always \texttt{Yes} or \texttt{No}) by setting the max new tokens to 1. This makes the output deterministic and ensures any differences we observe are due to changes in prompting or training. 

\subsubsection{Few-shot}
We then experimented with both LLMs using a Few-shot approach. To generate examples (the ``shots'') we selected 4 motif expressions, each for a different motif, one from each class. We also added negative examples paired with each example, selected at random. Each triplet of motif, positive example, and negative example was grouped together in the prompt, as shown in Appendix~\ref{sec:prompts}. As in the Zero-shot approach, we forced the models to output a deterministic single token (which was always \texttt{Yes} or \texttt{No}) by setting the temperature to 0 and the max new tokens to 1. 

\subsubsection{LoRA Fine-Tuning}
Finally, we fine-tuned the two LLMs using Low-Rank Adaptation (LoRA), a parameter-efficient method~\citep{hu2022lora}. The goal was to improve classification of whether a sentence contains a motif or not. We fine-tuned the LLMs using the same dataset we used in finetuning embedding models. Each training example consists of motif-sentence pairs as input, target output (Yes or No) only. We set epochs to 5 and learning rate to $10^{-4}$. 
 
We evaluated each of the three approaches like a normal Yes or No classifier using the test set. The evaluation is straightforward by seeing the responses of the models and comparing them with the gold labels that the annotators have provided.  We report precision, recall, and $F_1$ scores in Table~\ref{tab:results}.

\section{Discussion and Future Work}
\label{sec:discussion}

First, the good annotation agreement of our annotation study demonstrates that identifying whether a motif is present in text is an achievable task for people. Further, our system performance aligns with our complexity analysis, where we predict that certain motifs and motif expressions are more or less challenging due to their structural and expression complexity. We found that most motifs have simple conceptual structure, and most motif expressions were simple, meaning that system performance is dominated by performance on simple motifs simply expressed. 

Regarding performance, we found that while retrieve-and-rerank performed the worst in the long run, it worked well for simple motifs, and served as a starting point. In our implementation, we fixed the number of sentences returned by retrieve-and-rerank, which necessarily limits its performance in cases where a motif has more than 100 motif expressions across the corpus. Despite this limitation, the retrieve-and-rerank allowed us to quickly generate a new annotated dataset used to explore other methods embeddings and generative approaches. 

The six embedding models achieved an average $F_1$ score of 0.57. In the Zero-shot setup, Mistral-7B-Instruct-v0.3 performs similarly to the embedding models, despite its larger size (7B parameters), achieving an overall $F_1$ score of 0.61. However, with in the Few-shots setup, Mistral's $F_1$ score jumped by 17 points of $F_1$, to 0.78. The best performing models were the fine-tuned LLMs, achieving 0.81 (Mistral 7B) and 0.85 (Llama 3.1) $F_1$. Unlike Mistral, Llama's $F_1$ scores were relatively close under the Zero-shot and Few-shot setups. On average, when not provided with examples of how motifs appear in text, the LLMs achieved an overall $F_1$ score of 0.67, which shows that despite LLMs capabilities, this is still a challenging task for them.


For future work, we suspect that more fine-tuning will result in additional performance increases. Second, it seems promising to experiment with fully generative approaches, in particular, trying LLMs with longer context windows such as GPT-4o~\cite[128k;][]{openai2024gpt4o} or Gemini 2.0 Pro~\cite[2M tokens;][]{gemini2pro} at the same time for better complex motif detection that spans across multiple sentences. This would help avoid our current limitation in motif extraction, namely that we eliminate context and limit ourselves to a single sentence. Because of this complex motifs that find a cross span of text will not be fully detectable (this is admittedly a small fraction of expressions, however).  Another potential path is Retrieval-Augmented-Generation~\cite[RAG;][]{lewis2020retrieval}, which might allow us to completely avoid fine-tuned embeddings. Under a RAG approach, we can add more description to the prompt drawn from additional resources, such as the \textit{The Arabian Nights Encyclopedia}~\cite{marzolph2004arabian}, which contains concise descriptions of many of the motifs in the index. Finally, while we have collected many annotated examples, (200 motifs with 2,670 positive examples and 55,779 negative examples), we believe that continuing to expand the annotated data will allow development of even more capable models.


\section{Contributions}
\label{sec:contributions}

We analyzed motifs based on their conceptual complexity, and motif expressions based on their expression complexity, resulting in a four-way classification. We used a single text (\textit{The Arabian Nights}) paired with a single motif index (El-Shamy's \textit{A Motif Index of The Thousand and One Nights}) as our raw data. The motif index refers to each motif as a classic edition of the stories that are difficult to read and understand in today’s language due to their outdated terminology. Therefore, we aligned Burton’s translation to modern editions of the Arabian Nights narrative with high accuracy using two approaches. Using a machine-in-the-loop annotation strategy, we annotated 58,450 sentences for the presence or absence of various motifs (covering 200 motifs total), resulting in 2,670 labeled positive examples with an overall agreement of $0.72\kappa$. For the indexing task, the most effective models were LLM approaches using fine-tuned Mistral and Llama3, achieving an overall $F_1$ performance of 0.81 and 0.85, respectively. The study shows that this task is still challenging for state-of-the-art LLMs when not fine-tuned or given examples (i.e., no prior knowledge about motifs and how they appear in text). We identified possible next steps.




\section{Conflicts of interest}
The authors declare that they have no competing interests.

\section{Funding}
This work was supported in part by a Saudi Arabian Cultural Mission Fellowship to Ibrahim Alyami from the College of Computer Science and Information Systems at Najran University, Saudi Arabia [grant number 443-16-40].






\section{Data availability}
Upon acceptance of the paper, the code and data for reproducing this study will be deposited in the FIU DataVerse institutional electronic repository found at \url{https://dataverse.fiu.edu/}, which will provide permanent archiving with a DOI, which we will include in the final copy.


\section*{AI Disclosure Statement}
We used ChatGPT versions 4 and 5 for coding assistance only. All research decisions, annotations, analyses, evaluations, writing, and everything else were performed entirely by the authors.

\section{Author contributions statement}
I.A. contributed to: Conceptualization, Methodology, Software, Validation, Investigation, Data Curation, Writing - Original Draft, and Writing - Review \& Editing. M.A.F. contributed to: Conceptualization, Methodology, Resources, Writing - Review \& Editing, and Supervision.

\section{Acknowledgments}
We thank our annotators, Diego Ramos, Akshai Srinivasan, and Alain De Armas.




\bibliographystyle{unsrtnat}
\bibliography{paper}


\clearpage

\begin{appendices}

\section{Appendix A: Annotation Guideline}
\label{sec:annotationguidelines}
Text annotation can be a tedious and, at times, challenging task, depending on the subject and complexity of the data. It is crucial for the annotator to stay focused and fully immersed in the task at hand. Allowing external factors to influence the process can have adverse effects on the project’s outcome. While having a large amount of data and several annotators may offset this risk, it is important to maintain clean and accurate annotation results, as these are used to train models. Poor data quality leads to poor model performance, ultimately affecting the accuracy of predictions.

These annotation guidelines are built based on a specific scenario but can be generalized to other cases. In this example, we are working with motifs and sentences that may contain those motifs. The task is to label sentences as True Positive (TP) when the motif is present and False Positive (FP) when it is not. The sentences and motifs fall into four categories: Simple Expression, Complex Expression, Simple Structure, and Complex Structure. To illustrate, we will use the motif "Love at first sight" with two example sentences. The TP sentence is: “She uncovered her face, looked at him, and discovering him to be a handsome young man, she fell in love with him at first sight.” The FP sentence is: “The girl greeted me in the loveliest and sweetest voice I had ever heard, and when she uncovered her face, I saw that she was as radiant as the moon.” Although the motif in this case is simple, more complex scenarios may arise, so this is just for demonstration purposes.

The first step in the annotation process is to analyze the motif, “Love at first sight.” It consists of three key elements: “love,” “first,” and “sight.” When reviewing data, all three elements must be present for the sentence to be labeled as TP. In the first sentence, love is clearly described, and the context indicates it is happening for the first time upon seeing the other person, satisfying all three elements, so it is labeled TP. However, in the second sentence, while "sight" is present, there is no indication of it being the first time, nor is love involved, so it is labeled FP.

It is essential to ensure every element of the motif is present when annotating. As an annotator, you might feel tempted to label something as TP because you want it to be, rather than because it truly meets the criteria. This bias often occurs after a long string of FPs. Recognizing and avoiding this bias is important, as mislabeled data can pollute the dataset, leading to inaccurate model training.

Annotation can be mentally and physically exhausting, leading to mistakes or rushed work. To mitigate this, I recommend working on one-hour blocks with 15-minute breaks. While individual endurance varies, this 1/15 pattern is a good general suggestion to avoid fatigue-related errors.

In this specific setting, the data is drawn from an index (motif) and a set of sentences. These sentences are isolated from their broader context, which can cause difficulties. For example, proper names may appear without their associated titles, leading to potential mislabeling. If a motif is related to a king and the sentence mentions “Ibrahim,” without knowing that Ibrahim is a king, a TP could mistakenly be labeled FP, and vice versa. To address this, we propose applying a contextualized annotations algorithm, which provides annotators with the necessary context to make accurate decisions.

Working in teams can also simplify the annotation process, even though it may not always be feasible. When teams are in place, there are additional practices that can significantly enhance the quality of the output. Annotators might feel uncertain about particular data points, and in such cases, it is vital to flag those points for team discussion. Once individual members have completed their portions, team meetings should be held to review these flagged points and resolve any discrepancies. Collaborative discussions allow the team to finalize decisions on ambiguous data and ensure that all annotations are as consistent and accurate as possible.

In summary, when annotating data, it is important to follow these guidelines to ensure cleaner and more accurate results: Perform element identification, be honest by annotating with your head and not your heart, avoid rushing, always utilize context, and, when possible, collaborate effectively within a team. By adhering to these practices, annotators can produce high-quality data that ultimately leads to better model performance and more reliable results. The goal is to minimize errors and maximize the integrity of the annotated data, ensuring that the resulting models can perform at their best.

\section{Appendix B: Zero and Few-Shot Prompt}
\label{sec:prompts}
System message is the same for zero-shot and few-shot:
\begin{quote}
\tt System: You are a strict classifier. You must answer with ONLY one token: Yes or No.
\end{quote}

\subsection{Zero-shot prompt}
\begin{quote}
\tt Task: Decide if the motif is present in the sentence.\\ 
Rules: Answer ONLY ``Yes'' or ``No''. \\ 
Do not explain.\\
Motif:  <MOTIF>\\
Sentence:  <Sentence>
\end{quote}

\subsection{Few-shot prompt}
\begin{quote}
\tt Examples:\\
Motif: Mermaid \\
Sentence: While he was doing this the sea became disturbed and out from it came mermaids the sea’s daughters each carrying in her hand a jewel gleaming like a lamp.\\
Answer: Yes\\

Motif: Mermaid\\
Sentence: She can look as she stands and she will not be long.\\
Answer: No\\

Motif: Barber as know-all expert\\
Sentence: The attendant then started to shave Dau’ alMakan’s head after which he and the furnace man bathed.\\
Answer: Yes\\

Motif: Barber as know-all expert\\
Sentence: He is the expert in the field they said a very wealthy man and a skilled craftsman.\\
Answer: No\\

Motif: Blind promise of immunity from punishment. Person of authority (king, queen, father, etc.) grants request for safety for culprit before learning nature of offense \\
Sentence: My brother said ‘I want a guarantee of immunity’ at which the wali gave him the kerchief that was a sign of this.\\
Answer: Yes\\

Motif: Blind promise of immunity from punishment. Person of authority (king, queen, father, etc.) grants request for safety for culprit before learning nature of offense\\
Sentence: I am the shaikh of a monastery and under my control and authority are forty dervishes.\\
Answer: No\\

Motif: Adam created from clay (mud), mud foam (zabad), foam from sea, sea from darkness, darkness from bull, bull from whale, whale from rock, rock from ruby (gem), ruby from water, water from [God's] Omnipotence (al-Qudrah)\\
Sentence: The darkness itself was created from light, which was created from a fish, with the fish being created from a rock, the rock from a ruby, the ruby from water and the water from the power of God, as He said, Almighty is He: ‘When He wishes for something, He commands 'Be' and it is.\\
Answer: Yes

Motif: Adam created from clay (mud), mud foam (zabad), foam from sea, sea from darkness, darkness from bull, bull from whale, whale from rock, rock from ruby (gem), ruby from water, water from [God's] Omnipotence (al-Qudrah)
Sentence: These are water earth light darkness and fruits’ she answered.\\
Answer: No\\ 

Task: Decide if the motif is present in the sentence.\\ 
Rules: Answer ONLY ``Yes'' or ``No''. \\ 
Do not explain.\\
Motif:  <MOTIF>\\
Sentence:  <Sentence>\\
Answer:

\end{quote}

\section{Appendix C: List of Motifs}
\label{sec: appendix}
The rest of the paper shows the list of motifs that we have used in our approaches.\\
\begin{table}[!ht]
\centering
\caption{List of Motifs}
\label{tab:motifs}
\begin{tabular}{|c|p{4.7cm}|c|c|c|c|}
\hline
\textbf{Motif ID} & \textbf{Motif} & \textbf{+Examples} & \textbf{-Examples} & \textbf{Expression} & \textbf{Conceptual} \\
\hline
J0 & {Acquisition and possession of wisdom} & 43 & 258 & Simple\&Complex & Simple \\
\hline
A630.1.1 & {Adam created from clay (mud), mud foam (zabad), foam from sea, sea from darkness, darkness from bull, bull from whale, whale from rock, rock from ruby (gem), ruby from water, water from [God's] Omnipotence (al-Qudrah)} & 2 & 192 & Simple\&Complex & Complex \\
\hline
F1041.15.1 & {Addiction as an illness. Adverse effects of excessive consumption of commodity or service (e.g., food, drink, drug, or sex, entertainment, etc.)} & 4 & 193 & Simple\&Complex & Complex \\
\hline
K1514.4.1.1 & {Adulteress refuses to admit husband under pretense that he is a stranger. [Husband forgot password]} & 1 & 237 & Simple\&Complex & Complex \\
\hline
K1966 & {Alchemist} & 11 & 188 & Simple\&Complex & Simple \\
\hline
A1174.7 & {Angel of nights and day} & 1 & 279 & Simple\&Complex & Simple \\
\hline
P551 & {Army} & 133 & 176 & Simple\&Complex & Simple \\
\hline
P429.1 & {Astronomer} & 24 & 161 & Simple\&Complex & Simple \\
\hline
P343.1.1 & {Auctioneer's} & 52 & 150 & Simple\&Complex & Simple \\
\hline
D1620.1.2 & {Automatic doll} & 3 & 535 & Simple\&Complex & Simple \\
\hline
Q431.5.1 & {Banishment for attempted seduction} & 3 & 313 & Simple\&Complex & Simple \\
\hline
P778 & {Bankruptcy} & 34 & 166 & Simple\&Complex & Simple \\
\hline
P446 & {Barber} & 95 & 38 & Simple\&Complex & Simple \\
\hline
X252.3.3 & {Barber as know-all expert} & 50 & 315 & Simple\&Complex & Simple \\
\hline
B611.1 & {Bear paramour} & 6 & 476 & Simple\&Complex & Simple \\
\hline
F531.1.7.2 & {Black man} & 68 & 268 & Simple\&Complex & Simple \\
\hline
P446 & {Barber} & 95 & 38 & Simple\&Complex & Simple \\
\hline
X252.3.3 & {Barber as know-all expert} & 50 & 315 & Simple\&Complex & Simple \\
\hline
D1707.3 & {Blessed animals} & 19 & 429 & Simple\&Complex & Simple \\
\hline
D1707.2 & {Blessed bodily organ (limb)} & 4 & 293 & Simple\&Complex & Simple \\
\hline
D1707.3 & {Blessed animals} & 19 & 429 & Simple\&Complex & Simple \\
\hline
D1707.2 & {Blessed bodily organ (limb)} & 4 & 293 & Simple\&Complex & Simple \\
\hline
D1707 & {Blessed objects} & 16 & 435 & Simple\&Complex & Simple \\
\hline
\end{tabular}
\end{table}
\clearpage

\begin{table}[!ht]
\centering
\caption{List of Motifs}
\begin{tabular}{|c|p{4.8cm}|c|c|c|c|}
\hline
\textbf{Motif ID} & \textbf{Motif} & \textbf{+Examples} & \textbf{-Examples} & \textbf{Expression} & \textbf{Conceptual} \\
\hline
D1707.8 & {Blessed places} & 45 & 389 & Simple\&Complex & Simple \\
\hline
M224 & {Blind promise of immunity from punishment. Person of authority (king, queen, father, etc.) grants request for safety for culprit before learning nature of offense} & 2 & 214 & Simple\&Complex & Complex \\
\hline
T610.2.4 & {Boy's puberty} & 12 & 297 & Simple\&Complex & Simple \\
\hline
T415 & {Brother-sister incest} & 3 & 253 & Simple\&Complex & Simple \\
\hline
P448 & {Butcher} & 40 & 147 & Simple\&Complex & Simple \\
\hline
K199.2.1 & {Buying an item for “Whatever [price] you may say.” Trickster names impossible price (e.g., bushel of fleas of which half are males and half females-or the like)} & 9 & 177 & Simple\&Complex & Complex \\
\hline
Z1941.1.1 & {Camel symbolism} & 4 & 176 & Simple\&Complex & Simple \\
\hline
P456 & {Carpenter} & 36 & 138 & Simple\&Complex & Simple \\
\hline
V400 & {Charity} & 32 & 164 & Simple\&Complex & Simple \\
\hline
K92.4.1 & {Chess game won by distracting opponent's attention: girl makes seductive gestures (motions) that disorient her male opponent} & 5 & 184 & Simple\&Complex & Complex \\
\hline
P453.0.1 & {Cobbler} & 9 & 184 & Simple\&Complex & Simple \\
\hline
J1141.4 & {Confession induced by bringing an unjust action against accused. False message to thief's wife to send the stolen jewels case bribe to the judge. She does} & 4 & 202 & Simple\&Complex & Complex \\
\hline
P17.15 & {Conflict (war) over kingship} & 11 & 282 & Simple\&Complex & Simple \\
\hline
A5.5 & {Creation of the universe for the sake of a certain (sacred) person -(e.g., Abraham, Mohammed, Zoroaster/Zardusht)} & 1 & 191 & Simple\&Complex & Complex \\
\hline
N443 & {Dangerous secret(s) learned} & 8 & 296 & Simple\&Complex & Simple \\
\hline
R154.3 & {Daughter rescues father} & 2 & 303 & Simple\&Complex & Simple \\
\hline
E350.1 & {Dead hospitable person causes guest's animal to be slaughtered for food and then compensates guests for slaughtered animal (usually by providing a substitute)} & 2 & 241 & Simple\&Complex & Complex \\
\hline
T160.0.1 & {Defloration} & 4 & 199 & Simple\&Complex & Simple \\
\hline
P13.5.2.1 & {Disempowerment of ruler procedures (dethroning, impeaching)} & 11 & 196 & Simple\&Complex & Complex \\
\hline
B256.4 & {Domesticated wolves} & 6 & 405 & Simple\&Complex & Simple \\
\hline
J21.55.5 & {Don't commit an injustice} & 10 & 270 & Simple\&Complex & Simple \\
\hline
J21.55.1 & {Don't drink liquor} & 7 & 287 & Simple\&Complex & Simple \\
\hline
D1707.8 & {Druid as magician} & 40 & 283 & Simple\&Complex & Simple \\
\hline
P196.1.2.1 & {Drunkard strikes parent(s)} & 1 & 328 & Simple\&Complex & Simple \\
\hline
A844.2 & {Earth supported by bull} & 1 & 263 & Simple\&Complex & Simple \\
\hline
A844.3.1 & {Earth supported by whale} & 1 & 297 & Simple\&Complex & Simple \\
\hline

\end{tabular}
\end{table}
\clearpage

\begin{table}[!ht]
\centering
\begin{tabular}{|c|p{4.8cm}|c|c|c|c|}
\hline
A2921.1 & {Eblis: born as one of the fourteen children of Khalit and Malit. He disobeyed his father by refusing to marry one of his seven twin-sisters, and was transformed into worm (which became Eblis)} & 2 & 194 & Simple\&Complex & Complex \\
\hline
T320 & {Escape from undesired lover} & 2 & 190 & Simple\&Complex & Simple \\
\hline
T33.1 & {Faithless woman transforms lovers (husbands) to animals (birds) and keeps them as pets} & 8 & 188 & Simple\&Complex & Complex \\
\hline
F969.7 & {Famine} & 41 & 156 & Simple\&Complex & Simple \\
\hline
P140.0.3 & {Favor (pardon) won through intercession of person of influence} & 29 & 224 & Simple\&Complex & Complex \\
\hline
L51 & {Favorite youngest daughter} & 1 & 280 & Simple\&Complex & Simple \\
\textbf{Motif ID} & \textbf{Motif} & \textbf{+Examples} & \textbf{-Examples} & \textbf{Expression} & \textbf{Conceptual} \\
\hline
T187.0.2 & {Female's coition posture compared to supplication posture (pleading with God); (I.e., being on her back with legs raised toward heaven)} & 1 & 195 & Simple\&Complex & Simple \\
\hline
P413 & {Ferryman} & 15 & 184 & Simple\&Complex & Simple \\
\hline
L313 & {Few overcome numerous} & 13 & 292 & Simple\&Complex & Simple \\
\hline
B3.2 & {Fire-breathing viper} & 6 & 417 & Simple\&Complex & Simple \\
\hline
P419 & {Fisher} & 59 & 141 & Simple\&Complex & Simple \\
\hline
A102.1.2.1 & {Five things known only to God: Time of End of World (al-sa'ah), when and where rain will occur, [gender of] what is in wombs, one's future earnings, place of one's death} & 2 & 239 & Simple\&Complex & Complex \\
\hline
B41.2 & {Flying horse} & 13 & 369 & Simple\&Complex & Simple \\
\hline
P191.2 & {Foreigner (strangers) required to provide information on his social status and profession (craft, trade). (Usually at city gate)} & 12 & 178 & Simple\&Complex & Complex \\
\hline
N9.1 & {Gambler loses everything} & 10 & 309 & Simple\&Complex & Simple \\
\hline
F541.12.1 & {Gazelle's eyes} & 6 & 197 & Simple\&Complex & Simple \\
\hline
B873.4 & {Giant ant} & 7 & 388 & Simple\&Complex & Simple \\
\hline
B875.2 & {Giant crocodile} & 3 & 406 & Simple\&Complex & Simple \\
\hline
B873.2.10 & {Giant rhinoceros} & 4 & 374 & Simple\&Complex & Simple \\
\hline
B874.3 & {Giant whale} & 5 & 401 & Simple\&Complex & Simple \\
\hline
T39.1.4 & {Girl (woman) confesses to sexual offense (unchastity, infidelity) so as to save her sweetheart from punishment for theft. (His hand was about to be cut off for entering her home secretly)} & 1 & 198 & Simple\&Complex & Complex \\
\hline
T610.2.1 & {Girl's puberty} & 2 & 281 & Simple\&Complex & Simple \\
\hline
W160.1.1 & {Greedy (ambitious) person pursues large game (animal, fish) without regard to saftey: loses his life} & 4 & 182 & Simple\&Complex & Complex \\
\hline
G135 & {Habitat of giant ogres} & 1 & 293 & Simple\&Complex & Simple \\
\hline
B431.2 & {Helpful ape} & 22 & 338 & Simple\&Complex & Simple \\
\hline
B473 & {Helpful dolphin} & 2 & 480 & Simple\&Complex & Simple \\
\hline
B431.2 & {Helpful falcon} & 10 & 430 & Simple\&Complex & Simple \\
\hline
B431.2 & {Helpful lion} & 5 & 335 & Simple\&Complex & Simple \\
\hline
B431.2 & {Helpful monkey} & 32 & 413 & Simple\&Complex & Simple \\
\hline
\end{tabular}
\end{table}
\clearpage

\begin{table}[!ht]
\centering
\begin{tabular}{|c|p{4.7cm}|c|c|c|c|}
\hline

F1041.2 & {Horripilation. Hair rises on end in extraordinary fashion from joy, anger, or love.} & 2 & 194 & Simple\&Complex & Complex \\
\hline
P414 & {Hunter} & 65 & 104 & Simple & Simple \\
\hline
P12 & {Hunting a madness of kings} & 14 & 273 & Simple\&Complex & Simple \\
\hline
K1874.2.1 & {Husband driven insane with pseudo-doubles. His private possessions exhibited before him at his workplace, then quickly returned to his home before his arrival (usually via tunnel): he thinks he has seen a double and regrets having suspected his wife} & 4 & 228 & Simple\&Complex & Complex \\
\textbf{Motif ID} & \textbf{Motif} & \textbf{+Examples} & \textbf{-Examples} & \textbf{Expression} & \textbf{Conceptual} \\
\hline
 K1534.1 & {Husband made to sift dirt. Wife, whose food purchase was replaced with dirt and rocks during her sexual liaison with merchant, claims that she dropped the money on ground and that she brought the dirt home to search for it} & 6 & 201 & Simple\&Complex & Complex \\
\hline
R151 & {Husband rescues wife} & 1 & 298 & Simple\&Complex & Simple \\
P509.3 & {Immunity of ruler (caliph, king, president, etc.) from personal responsibility for mistakes. Tendency of subjects (narrators) to view ruler as balmeless (or as victim of bad advice)} & 1 & 532 & Simple\&Complex & Complex \\
\hline
P761 & {Inheritance} & 52 & 139 & Simple\&Complex & Simple \\
\hline
D1711.1.4 & {Jew as magician} & 10 & 367 & Simple\&Complex & Simple \\
\hline
F499.3.5.3.1 & {Jinn dwell in cemeteries} & 2 & 277 & Simple\&Complex & Simple \\
\hline
P180.3 & {Joint ownership of slave} & 1 & 288 & Simple\&Complex & Simple \\
\hline
W35 & {Justice} & 94 & 80 & Simple\&Complex & Simple \\
\hline
U10 & {Justice and injustice} & 17 & 154 & Simple\&Complex & Simple \\

P11.1.1 & {King chosen by lot} & 6 & 285 & Simple\&Complex & Simple \\
\hline
B242 & {King of birds} & 9 & 341 & Simple\&Complex & Simple \\
\hline
B242.2.1 & {King of crows} & 2 & 404 & Simple\&Complex & Simple \\
\hline
B241.2.2 & {King of monkeys} & 12 & 418 & Simple\&Complex & Simple \\
\hline
P14.26.1 & {King should aid friend (ally) who is being attacked} & 7 & 201 & Simple\&Complex & Complex \\
\hline
M203 & {King's promise irrevocable} & 4 & 337 & Simple\&Complex & Simple \\
\hline
B225.1 & {Kingdom of serpents} & 35 & 402 & Simple\&Complex & Simple \\
\hline
P14.12.0.1 & {Kings exchange presents (gifts)} & 26 & 279 & Simple\&Complex & Simple \\
\hline
N650 & {Life saved by accident} & 2 & 288 & Simple\&Complex & Simple \\
\hline
B15.7.18 & {Locus as hybrid of many (mighty) animals: horse's face, elephant's eyes, ox's neck, stag's antlers, lion's chest, scorpion's belly, eagle's wings, camel's thigh, ostrich's leg, viper's tale} & 16 & 220 & Simple\&Complex & Complex \\
\hline
T15 & {Love at first sight} & 30 & 312 & Simple\&Complex & Simple \\
\hline
Z97.7 & {Lover's (or brother-sister's) alliterative names mirroring each other (e.g., gods Nun-Nanoid, gods Amon-Agminate, Hamas-Hamadah, Sami-Samyah, etc.). “Phonetic bifurcation”} & 14 & 250 & Simple\&Complex & Complex \\
\hline

\end{tabular}
\end{table}
\clearpage

\begin{table}[!ht]
\centering
\begin{minipage}{3\columnwidth}
\begin{tabular}{|c|p{5.4cm}|c|c|c|c|}
\hline
D1266 & {Magic book} & 5 & 406 & Simple\&Complex & Simple \\
\hline
D1272 & {Magic circle} & 4 & 392 & Simple\&Complex & Simple \\
\hline
D1242.0.1 & {Magic elixir (nectar)} & 21 & 448 & Simple\&Complex & Simple \\
\hline
D2031 & {Magic illusion} & 23 & 401 & Simple\&Complex & Simple \\
\hline
D1071 & {Magic jewel (jewels)} & 13 & 324 & Simple\&Complex & Simple \\
\hline
D1254.2 & {Magic rod} & 24 & 376 & Simple\&Complex & Simple \\
\hline
D1960 & {Magic sleep} & 8 & 345 & Simple\&Complex & Simple \\
\hline
C1408.1 & {Magic sphere burns up country. By turning that part of the globe to the sun, one can make any place on earth burn up} & 2 & 186 & Simple\&Complex & Complex \\
\hline
D927 & {Magic spring} & 7 & 439 & Simple\&Complex & Simple \\
\hline
D1254 & {Magic staff} & 68 & 348 & Simple\&Complex & Simple \\
\hline
\textbf{Motif\#} & \textbf{Motif} & \textbf{Total TP} & \textbf{Total FP} & \textbf{Expression} & \textbf{Conceptual} \\
\hline
D1500.1.34 & {Magic writings heal} & 2 & 289 & Simple\&Complex & Simple \\
\hline
T610.1 & {Maturation (sexual)} & 18 & 191 & Simple\&Complex & Simple \\
\hline
B81 & {Mermaid} & 4 & 195 & Simple\&Complex & Simple \\
\hline
B82 & {Merman} & 33 & 139 & Simple\&Complex & Simple \\
\hline
D625 & {Monthly transformation} & 11 & 426 & Simple\&Complex & Simple \\
\hline
V235 & {Mortal visited by angel} & 15 & 240 & Simple\&Complex & Simple \\
\hline
Z117.6.1 & {Musical instrument speaks} & 2 & 185 & Simple\&Complex & Simple \\
\hline
C434 & {Names of dangerous things (animal, disease, murder, etc.) are not to be uttered at a person without the use of precautionary measures (e.g., “Distant one,” “Away from you”)} & 14 & 188 & Simple\&Complex & Complex \\
\hline
M114.1.3 & {Oath (vow) taken on Koran} & 2 & 283 & Simple\&Complex & Simple \\
\hline
M113.1 & {Oath taken on sword} & 1 & 288 & Simple\&Complex & Simple \\
\hline
N825.3 & {Old woman helper} & 18 & 273 & Simple\&Complex & Simple \\
\hline
N543.4 & {Only one person can overcome treasure's protective measures (rasad/incantations) and open it} & 9 & 224 & Simple\&Complex & Complex \\
\hline
A2921 & {Origin of Iblis} & 2 & 453 & Simple\&Complex & Simple \\
\hline
A2813 & {Origin of honey} & 2 & 338 & Simple\&Complex & Simple \\
\hline
A2812 & {Origin of musk} & 2 & 384 & Simple\&Complex & Simple \\
\hline
A2811 & {Origin of silk} & 1 & 305 & Simple\&Complex & Simple \\
\hline
A2814 & {Origin of spices} & 3 & 407 & Simple\&Complex & Simple \\
\hline
T603.1 & {Pampered son(s)} & 19 & 327 & Simple\&Complex & Simple \\
\hline
N393.1 & {Parasite (sponger) sneaks into a party of seemingly important men, accompanied by guards, thinking that they are being escorted to feast. They prove to be prisoners to be excuted} & 1 & 221 & Simple\&Complex & Complex \\
\hline
S22.1 & {Parricide to obtain kingship} & 2 & 302 & Simple\&Complex & Simple \\
\hline
K490.2 & {Performing prayers as excuse} & 5 & 290 & Simple\&Complex & Simple \\
\hline
K712.3.2 & {Person invited to attend religious ritual (visit saint, magic healing, etc.) but is taken to an isolated place where he is attacked (robbed, raped, or the like)} & 1 & 222 & Simple\&Complex & Complex \\
\hline
F669.1.1 & {Person so skilled in administration (managing others) that he can use a spider web as harness for unruly group} & 8 & 186 & Simple\&Complex & Complex \\
\hline
P424 & {Physician} & 14 & 184 & Simple\&Complex & Simple \\
\hline
\end{tabular}
\end{minipage}
\end{table}
\clearpage

\begin{table}[!ht]
\centering
\begin{minipage}{3\columnwidth}
\begin{tabular}{|c|p{4.8cm}|c|c|c|c|}
\hline
\textbf{Motif ID} & \textbf{Motif} & \textbf{+Examples} & \textbf{-Examples} & \textbf{Expression} & \textbf{Conceptual} \\
\hline
K2139 & {Poisonous medicine for king: envious man gives king's barber (doctor, etc.) a supposedly healing substance (actually poison), then he reports to the king that the barber plans to assassinate him with that ‘medicine’} & 4 & 188 & Simple\&Complex & Complex \\
\hline
P5.3 & {Possessions as indicators of social status} & 58 & 250 & Simple\&Complex & Complex \\
\hline
V59.0.1 & {Prayers answered especially when gates of sky (heavens) are open} & 11 & 185 & Simple\&Complex & Complex \\
\hline
F679.12.1 & {Precious stones (diamonds, emeralds, etc.) retrieved from bottom of inaccessible valley with the help of vultures. (Meat thrown from great heights into valley, stones adhere to meat, vultures carry meat along with stones to valley ridge where miners can collect them)} & 5 & 177 & Simple\&Complex & Complex \\
J1913.6 & {Price of an ‘inexpensive’ article (service) raised through mutually misunderstood gesture (murmur). Owner thinks buyer is ridiculing him by offering too high a price and makes gestures to that effect, but buyer thinks owner thinks offer is too low and raises his offer-(this happens repeatedly)} & 8 & 192 & Simple\&Complex & Complex \\
\hline
L400 & {Pride brought low} & 17 & 268 & Simple\&Complex & Simple \\
\hline
P402 & {Professional code of honor (rules of conduct for members of guild or profession; e.g., physicians, goldsmiths, clerics, water-carriers)} & 4 & 216 & Simple\&Complex & Complex \\
\hline
H709.1.1 & {Puzzle: Part of a flock of pigeons alighted on tree while another alighted on ground. The ones on the tree said to the ones on the ground “If one of you joined us on top your number becomes on third of all of us, but if one of us joined you on the ground your number becomes one half of all of us.” How many pigeons were in the flock? (12: 7 on tree, 5 on ground)} & 8 & 187 & Simple\&Complex & Complex \\
\hline
B244.1.1.1 & {Queen of vipers} & 30 & 421 & Simple\&Complex & Simple \\
\hline
H11.1.4 & {Recognition by characteristic handwriting} & 8 & 299 & Simple\&Complex & Simple \\
\hline
H11.1.4 & {Recognition by tracing ancestry} & 8 & 277 & Simple\&Complex & Simple \\
\hline
Q46 & {Reward for protecting fugitive} & 2 & 290 & Simple\&Complex & Simple \\
\hline
P793 & {Rivalry: social interactional process. Recurrent conscious competition between the same parties} & 40 & 354 & Simple\&Complex & Complex \\
\hline
B31.1.1 & {Roc's egg} & 12 & 451 & Simple\&Complex & Simple \\
\hline
B811.8 & {Sacred dog} & 2 & 420 & Simple\&Complex & Simple \\
\hline
B811.9 & {Sacred donkey (ass)} & 1 & 381 & Simple\&Complex & Simple \\
\hline
B811.10 & {Sacred mule} & 2 & 345 & Simple\&Complex & Simple \\
\hline
\end{tabular}
\end{minipage}
\end{table}
\clearpage

\begin{table}[!ht]
\centering
\begin{tabular}{|c|p{4.8cm}|c|c|c|c|}
\hline
\textbf{Motif ID} & \textbf{Motif} & \textbf{+Examples} & \textbf{-Examples} & \textbf{Expression} & \textbf{Conceptual} \\
\hline
B811.10.1 & {Sacred she-mule} & 1 & 635 & Simple\&Complex & Simple \\
\hline
B811.7 & {Sacred wolf} & 2 & 352 & Simple\&Complex & Simple \\
\hline
B71 & {Sea horse} & 4 & 416 & Simple\&Complex & Simple \\
\hline
B215.5  & {Serpent language} & 1 & 468 & Simple\&Complex & Simple \\
\hline
P234.0.3.1 & {Seven daughters} & 7 & 358 & Simple\&Complex & Simple \\
\hline
A874 & {Seven strata of earth} & 10 & 282 & Simple\&Complex & Simple \\
\hline
K301.2.1 & {Several brothers as robbers} & 1 & 286 & Simple\&Complex & Simple \\
\hline
E783.5.1 & {Severed head speaks} & 2 & 293 & Simple\&Complex & Simple \\
\hline
B754 & {Sexual habits of animals} & 4 & 295 & Simple\&Complex & Simple \\
\hline
P412 & {Shepherd} & 30 & 151 & Simple\&Complex & Simple \\
\hline
D1 & {Sihr (magic, sorcery): controlling (coercing, harnessing) the supernatural and the natural by means of supernatural agents other than God and His powers} & 23 & 162 & Simple\&Complex & Complex \\
\hline
P428.0.1 & {Singers} & 44 & 151 & Simple\&Complex & Simple \\
\hline
T52.0.6.1.2 & {Slave-girl gives owner-to-be money with which to buy her.(She is infatuated with him, while he is impoverished and cannot afford the high price)} & 13 & 231 & Simple\&Complex & Complex \\

B211.6.1 & {Speaking snake (serpent)} & 23 & 292 & Simple\&Complex & Simple \\
\hline
A64 & {Spying satan(s): devil attempt(s) to learn heavenly secrets by eavesdropping on sky-worlds} & 2 & 191 & Simple\&Complex & Complex \\
\hline
S38.1 & {Stepson kills stepmother} & 1 & 322 & Simple\&Complex & Simple \\
\hline
U5.1 & {Successor falls short} & 4 & 291 & Simple\&Complex & Simple \\
\hline
D361.1 & {Swan Maiden} & 2 & 506 & Simple\&Complex & Simple \\
\hline
C60 & {Tabu: violators of ablution-state (wudu: being ritually clean)-ritual contaminants (nagasah): acts and objects that defile, or cause ritual uncleanliness and becoming unfit to perform certain religious rituals} & 10 & 194 & Simple\&Complex & Complex \\
\hline
T404.4 & {Temptress seeks to seduce man} & 18 & 246 & Simple\&Complex & Simple \\
\hline
H509.4 & {Test of poetic ability} & 8 & 292 & Simple\&Complex & Simple \\
\hline
H808 & {The cycle of the sinful-legitimate amorous regard (glance). A man looked at someone else's slave-girl in the morning (sinful), at noon he purchased her for himself (became legitimate), in the afternoon he freed her (became sinful), at sunset he married her (became legitimate), in late evening he divorced her (became sinful), in the morning he restored her (became legitimate)} & 6 & 187 & Simple\&Complex & Complex \\
\hline
F567.5.1 & {The desert-loner: self-banished man lives alone in the desert. (Sometimes accompanied by only one favorite person)} & 1 & 189 & Simple\&Complex & Complex \\
\hline
\end{tabular}
\end{table}
\clearpage

\begin{table}[!ht]
\centering
\begin{tabular}{|c|p{4.6cm}|c|c|c|c|}
\hline
\textbf{Motif ID} & \textbf{Motif} & \textbf{+Examples} & \textbf{-Examples} & \textbf{Expression} & \textbf{Conceptual} \\
\hline
W290.2 & {The good wife (woman): no divorce, no remarriage, no desertion due to vanity (batar), no disagreement with husband, no quittuing of husband's home to parent's due to unhappiness (gadbanah)} & 10 & 197 & Simple\&Complex & Complex \\
\hline
K2151.2 & {The seemingly dead is abandoned in street and made to look as if leaning against wall: one unsuspecting passer-by after another tries to awaken him and thinks that he caused his death} & 3 & 213 & Simple\&Complex & Complex \\
\hline
K366.1.5 & {Theft by trained monkey} & 1 & 266 & Simple\&Complex & Simple \\
\hline
B778.5 & {Thieving bird} & 7 & 302 & Simple & Simple \\
\hline
P234.0.3.1 & {Three daughters} & 5 & 380 & Simple\&Complex & Simple \\
\hline
M412 & {Time of giving curse} & 3 & 299 & Simple\&Complex & Simple \\
\hline
F69 & {Tour of sky-worlds} & 1 & 312 & Simple\&Complex & Simple \\
\hline
D113.2 & {Transformation: man to bear} & 4 & 290 & Simple\&Complex & Simple \\
\hline
N534 & {Treasure discovered by accident} & 2 & 283 & Simple\&Complex & Simple \\
\hline
P790.1.2.1 & {Trellis (ululation) of joy. (Typically voiced by women at a joyous occasion such as wedding, pilgrimage, winning at law court, release from prison, etc.)} & 2 & 196 & Simple\&Complex & Complex \\
\hline
K992.1 & {Trickster seeking revenge on blind men feigns blindness and claims that a beating by the police has ‘cured’ (reformed) him; he recommends the same ‘cure’ for his blind adversaries} & 13 & 232 & Simple\&Complex & Complex \\
\hline
P500.0.1 & {Types of rule (government): hereditary, election, selection, revolutionary-council, choice by lot (etc.)} & 24 & 193 & Simple\&Complex & Complex \\
\hline
Z13.11.1 & {Uncertainty about accuracy of truthful report: “And God knows best”,“And God is Omniscient” (or the like)} & 24 & 160 & Simple\&Complex & Complex \\
\hline
F956.7.7.2 & {Venting frustration (expression sorrow) by causing pain to oneself (hitting own head, slapping own face, biting own finger, or the like)} & 22 & 181 & Simple\&Complex & Complex \\
\hline
V515.1.1 & {Vision of chairs (thrones) in heaven.  Chairs of gold, silver, crystal (glass) assigned to saints according to merit} & 2 & 179 & Simple\&Complex & Complex \\
\hline
B143.1.3 & {Warning parrot} & 3 & 421 & Simple\&Complex & Simple \\
\hline
K251.7.1 & {Weighing eyes to see whether they are equal in value: “An eye for an eye.” The one-eyed accuser declines the test: he will be blinded while the accused would be left with one eye} & 9 & 193 & Simple\&Complex & Complex \\
\hline
\end{tabular}
\end{table}
\clearpage

\begin{table}[!ht]
\centering
\begin{tabular}{|c|p{4.6cm}|c|c|c|c|}
\hline
\textbf{Motif ID} & \textbf{Motif} & \textbf{+Examples} & \textbf{-Examples} & \textbf{Expression} & \textbf{Conceptual} \\
\hline
B731.11.1 & {White viper} & 12 & 373 & Simple\&Complex & Simple \\
\hline
R152 & {Wife rescues husband} & 5 & 293 & Simple\&Complex & Simple \\
\hline
T197 & {Wife's relatives (father, brother, etc.) force her divorce from husband without her consent} & 5 & 190 & Simple\&Complex & Complex \\
\hline
B3.3.1 & {Winged viper} & 2 & 429 & Simple\&Complex & Simple \\
\hline
B121.7 & {Wise tortoise} & 7 & 471 & Simple\&Complex & Simple \\
\hline
D658.3.4 & {Women (queen) transforms self to bird (animal) in order to copulate with men she had bewitched into birds (animals)} & 8 & 179 & Simple\&Complex & Complex \\
\hline
V1.11 & {Worship of idols} & 58 & 132 & Simple\&Complex & Simple \\
\hline
K1514.4.1.1 & {Would-be adulterer husband beaten by his would-be adulteress wife. Procuress brings man to woman, he proves to be her husband: wife beats him pretending that she was testing his fidelity} & 5 & 207 & Simple\&Complex & Complex \\
\hline
A661.0.1.1.5 & {‘Door of Deeds’: from Earth to heaven. A person's perodainted deeds ascend to heaven and enter via that door; it is shut when that person dies} & 7 & 186 & Simple\&Complex & Complex \\
\hline
A661.0.1.1.4 & {‘Door of Livelihood’: from heavens to Earth. A creature's perodainted sustenance is sent down from heaven via that door; it is shut when that creature's lifetime expires} & 4 & 239 & Simple\&Complex & Complex \\
\hline
\end{tabular}
\end{table}
\clearpage

\end{appendices}

\end{document}